\pdfoutput=1

\documentclass[10pt,twocolumn,letterpaper]{article}

\usepackage[pagenumbers]{cvpr} % To force page numbers, e.g. for an arXiv version

\usepackage[accsupp]{axessibility}  % Improves PDF readability for those with disabilities.

\definecolor{cvprblue}{rgb}{0.21,0.49,0.74}
\definecolor{markred}{rgb}{0.80,0.00,0.00}
\definecolor{deepfashion}{rgb}{0.80,0.10,0.10}
\definecolor{cars}{rgb}{0.90,0.32,0.09}
\definecolor{products}{rgb}{0.07,0.35,0.75}
\definecolor{landmarks}{rgb}{0.07,0.46,0.28}
\definecolor{learnable}{rgb}{0.92,0.40,0.18}
\definecolor{frozen}{rgb}{0.46,0.78,0.98}

\usepackage[pagebackref,breaklinks,colorlinks,allcolors=cvprblue]{hyperref}

\usepackage[dvipsnames]{xcolor}
\usepackage{color, colortbl}
\usepackage[utf8]{inputenc} % allow utf-8 input
\usepackage[T1]{fontenc}    % use 8-bit T1 fonts
\usepackage{url}            % simple URL typesetting
\usepackage{booktabs}       % professional-quality tables
\usepackage{amsfonts}       % blackboard math symbols
\usepackage{nicefrac}       % compact symbols for 1/2, etc.
\usepackage{mathtools}
\usepackage{graphicx}
\usepackage{wrapfig}
\usepackage{multirow}
\usepackage{adjustbox}
\usepackage{afterpage}
\usepackage{bbding}
\usepackage{cuted}
\usepackage{float}
\usepackage[most]{tcolorbox}

\def\paperID{38041}
\def\confName{CVPR}
\def\confYear{2026}

\title{
    \begin{minipage}{.06\textwidth}
        \centering
        \includegraphics[width=0.8\linewidth]{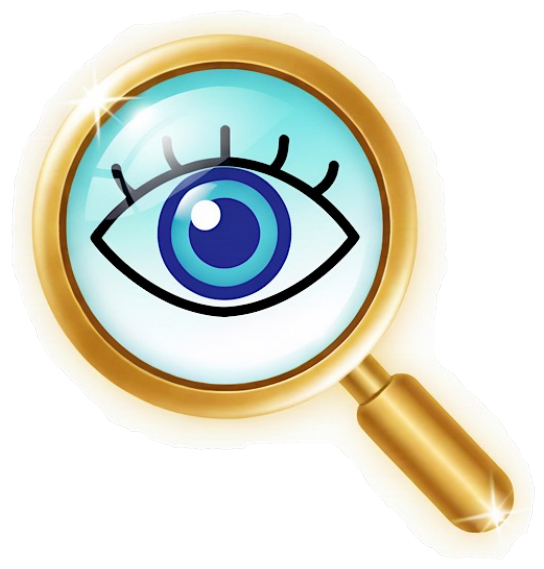} 
    \end{minipage}
        Beyond Semantic Search: Towards Referential Anchoring in \\
        Composed Image Retrieval
}

\makeatletter
\def\blfootnote{\gdef\@thefnmark{}\@footnotetext}
\makeatother

\author{
Yuxin~Yang$^{1,2}$  \quad  Yinan~Zhou$^{3,4}$  \quad  Yuxin~Chen$^{4}$  \quad  Ziqi~Zhang$^{1}$  \quad  Zongyang~Ma$^{1}$ \\
Chunfeng~Yuan$^{1,2,}$\thanks{Corresponding author.}  \quad  Bing~Li$^{1,2,5}$  \quad  Jun~Gao$^{6}$  \quad  Weiming~Hu$^{1,2,7}$ \vspace{1mm} \\
{ \normalsize
$^{1}$Institute of Automation, Chinese Academy of Sciences \quad
$^{2}$University of Chinese Academy of Sciences
} \\
{ \normalsize
$^{3}$Xi'an Jiaotong University \quad
$^{4}$Tencent Inc. \quad
$^{5}$PeopleAI Inc. \quad
$^{6}$HelloGroup Inc. \quad
$^{7}$ShanghaiTech University
} \\
{\tt\small
\{yangyuxin2023, mazongyang2020\}@ia.ac.cn, \{ziqi.zhang, cfyuan, bli, wmhu\}@nlpr.ia.ac.cn,
} \\
{\tt\small
zyn13572297710@stu.xjtu.edu.cn, uasonchen@tencent.com, gaojun55@gmail.com}
}

\begin{document}

\maketitle

\blfootnote{\textbf{Project page: \url{https://hahajun1101.github.io/OACIR/}}}

% \begin{strip}
%     \centering
%     \vspace{-15mm}
%         \includegraphics[width=\textwidth, trim=-2.5cm 1.0cm -2.5cm 0cm, clip]{Figures/OACIR_Task_Overview.pdf}
%     \captionof{figure}{Overview of the \textbf{O}bject-\textbf{A}nchored \textbf{C}omposed \textbf{I}mage \textbf{R}etrieval (\textbf{OACIR}) task and our \textbf{OACIRR} dataset.}
%     \label{fig:oacir_task_overview}
% \end{strip}

\begin{strip}
    \centering
    \vspace{-18mm}
        \includegraphics[width=\textwidth, trim=-2.5cm 1.0cm -2.5cm 0cm, clip]{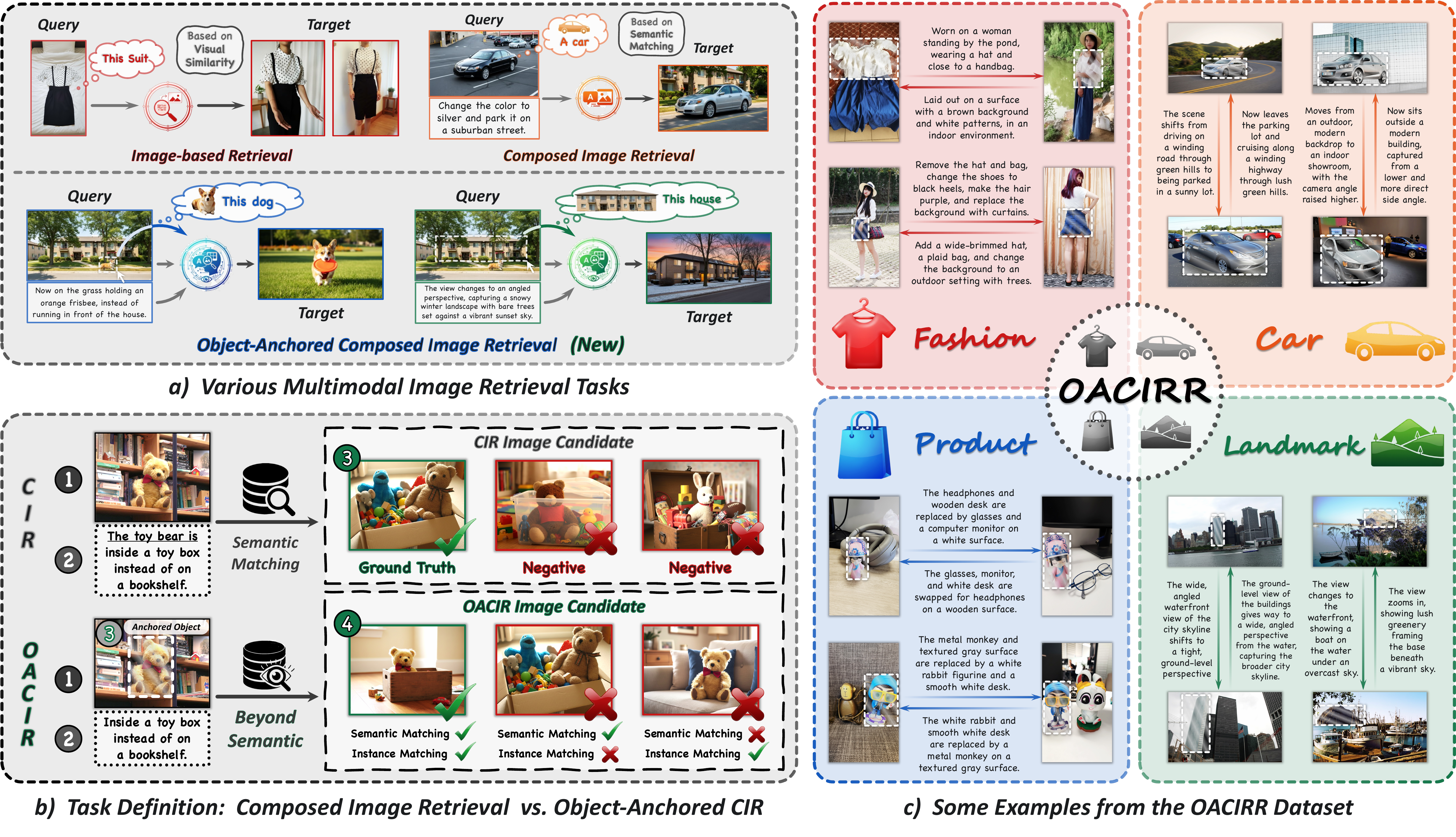}
    \vspace{-5mm}
    \captionof{figure}{Overview of the \textbf{O}bject-\textbf{A}nchored \textbf{C}omposed \textbf{I}mage \textbf{R}etrieval (\textbf{OACIR}) task and our \textbf{OACIRR} dataset.}
    \vspace{-1mm}
    \label{fig:oacir_task_overview}
\end{strip}

\begin{abstract}
\vspace{-6mm}

Composed Image Retrieval (CIR) has demonstrated significant potential by enabling flexible multimodal queries that combine a reference image and modification text.
However, CIR inherently prioritizes semantic matching, struggling to reliably retrieve a user-specified instance across contexts.
In practice, emphasizing concrete instance fidelity over broad semantics is often more consequential.
In this work, we propose \textbf{O}bject-\textbf{A}nchored \textbf{C}omposed \textbf{I}mage \textbf{R}etrieval (\textbf{OACIR}), a novel fine-grained retrieval task that mandates strict instance-level consistency.
To advance research on this task, we construct \textbf{OACIRR} (\textbf{OACIR} on \textbf{R}eal-world images), the first large-scale, multi-domain benchmark comprising over 160K quadruples and four challenging candidate galleries enriched with hard-negative instance distractors.
Each quadruple augments the compositional query with a bounding box that visually anchors the object in the reference image, providing a precise and flexible way to ensure instance preservation.
To address the OACIR task, we propose \textbf{AdaFocal}, a framework featuring a Context-Aware Attention Modulator that adaptively intensifies attention within the specified instance region, dynamically balancing focus between the anchored instance and the broader compositional context.
Extensive experiments demonstrate that \textbf{AdaFocal} substantially outperforms existing compositional retrieval models, particularly in maintaining instance-level fidelity, thereby establishing a robust baseline for this challenging task while opening new directions for more flexible, instance-aware retrieval systems.

\end{abstract}

\vspace{-3.5mm}
\section{Introduction}
\label{sec:introduction}

The paradigm of image retrieval has progressively evolved toward more flexible and user-oriented forms of interaction.
While traditional single-modal methods~\cite{visual_1, visual_2, textual_1, textual_2, textual_3} often struggle to express complex user intentions, Composed Image Retrieval (CIR)~\cite{tirg, artemis, clip4cir} has emerged as a powerful paradigm to address this limitation.
By combining a reference image with modification text, CIR leverages the synergy between visual and textual modalities to retrieve semantically aligned target images.
This capability has significantly broadened its applicability across diverse domains, including e-commerce and interactive search systems.

Despite its flexibility, the fundamental design of CIR prioritizes semantic matching over instance-level fidelity.
As illustrated in Figure~\ref{fig:oacir_task_overview}(a), the reference image in a conventional CIR query often serves as a coarse-grained visual anchor, defining the global visual scene or object category.
Consequently, the CIR model is tasked primarily with broad semantic integration, rendering the retrieval of a specific instance unreliable, particularly in the presence of visually similar distractors.
In many practical applications~\cite{cite_1, cite_2, cite_3}, including digital memory retrieval and long-term identity tracing, emphasizing concrete instance fidelity is often more critical than achieving broad semantic alignment.

In this work, we propose \textbf{O}bject-\textbf{A}nchored \textbf{C}omposed \textbf{I}mage \textbf{R}etrieval (\textbf{OACIR}), a novel fine-grained image retrieval task that mandates strict instance-level consistency.
As illustrated in Figure~\ref{fig:oacir_task_overview}(b), OACIR extends the conventional compositional query by incorporating an anchored instance.
The objective is to retrieve a target image that semantically satisfies the textual modification while strictly preserving the identical anchored instance.
Achieving this objective substantially advances compositional retrieval systems, enabling more flexible and expressive user interactions while improving reliability in real-world scenarios.
While offering these advantages, this powerful formulation also introduces two core challenges:
(1) \textbf{\textit{Compositional Reasoning}}: Requires the synthesis of three distinct information sources — the anchored instance, the global visual scene, and the textual modification — into a single coherent representation.
(2) \textbf{\textit{Fine-grained Discrimination}}: Requires distinguishing the exact anchored instance from a gallery enriched with visually and semantically similar distractors.

To advance research on this emergent task, we construct \textbf{OACIRR} (\textbf{OACIR} on \textbf{R}eal-world images), the first large-scale, multi-domain benchmark for OACIR.
As showcased in Figure~\ref{fig:oacir_task_overview}(c), \textbf{OACIRR} comprises a unified training set of 127K quadruples covering 2,647 instances, along with an extensive evaluation benchmark containing 33.4K queries across 1,238 instances from four diverse domains: Fashion, Car, Product, and Landmark.
The benchmark is enriched with over 26.6K curated distractor instances to form challenging galleries.
Collectively, \textbf{OACIRR} provides both a high-quality foundational dataset and a rigorous, comprehensive benchmark for the OACIR task.

To address the unique challenges of OACIR, we propose \textbf{\textit{AdaFocal}}, a simple yet effective framework that integrates a lightweight \textit{Context-Aware Attention Modulator (CAAM)}.
This module analyzes the multimodal query context to predict a modulation scalar, which is then used to adaptively intensifies visual attention on the anchored instance during feature fusion.
This mechanism achieves a dynamic balance between instance preservation and compositional reasoning.
Our extensive experiments validate that \textbf{\textit{AdaFocal}} substantially outperforms existing retrieval paradigms adapted for the OACIR task, demonstrating a pronounced advantage in maintaining instance-level fidelity.
These results not only establish \textbf{\textit{AdaFocal}} as a robust baseline but also underscore the significance of our benchmark in revealing the limitations of current semantic-level retrieval models.

In summary, the main contributions are as follows:

\begin{itemize}
\item
    We propose the novel \textbf{O}bject-\textbf{A}nchored \textbf{C}omposed \textbf{I}mage \textbf{R}etrieval (\textbf{OACIR}) task, which advances compositional retrieval beyond semantic matching by mandating strict instance-level consistency.
\item
    We construct \textbf{OACIRR}, a large-scale, multi-domain benchmark comprising over 160K real-world quadruples from 3.9K unique instances, and a challenging evaluation protocol tailored for rigorous instance-level assessment.
\item
    We propose \textbf{\textit{AdaFocal}}, an efficient framework that dynamically intensifies attention on the anchored instance region, providing a robust baseline for the OACIR task.
\end{itemize}

\begin{figure*}[t]
    \vspace{-2.5mm}
    \centering
        \includegraphics[width=0.98\linewidth]{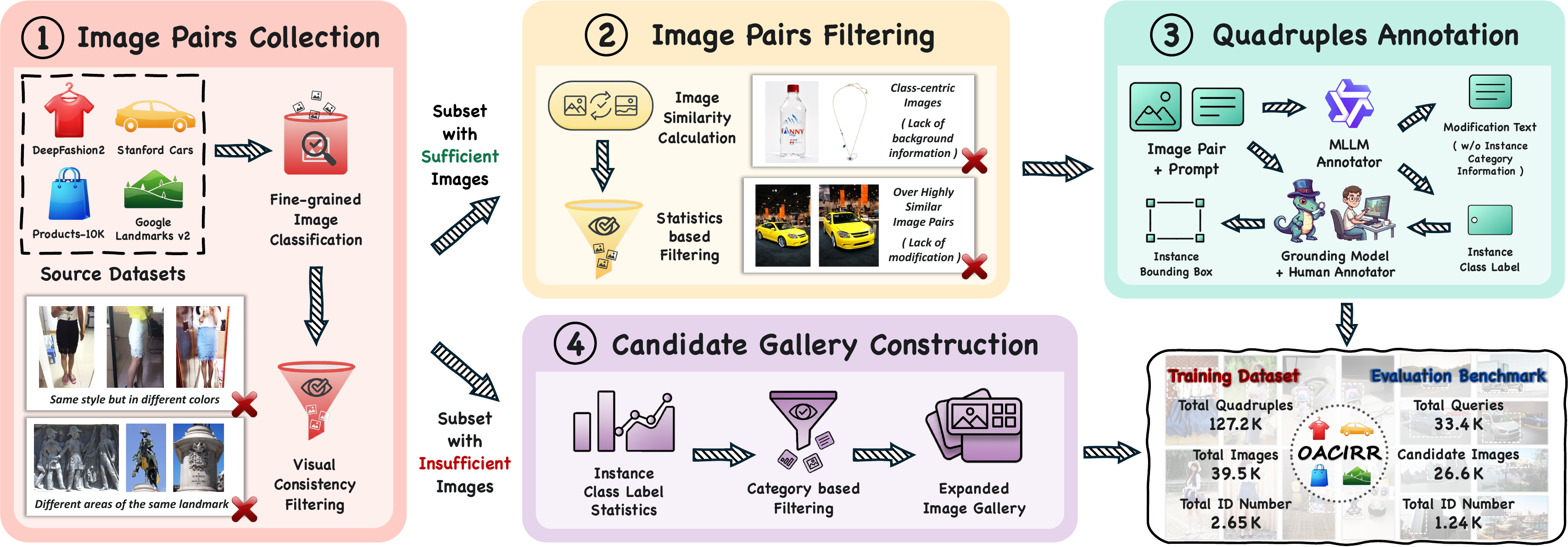}
    \caption{The multi-stage construction pipeline for the \textbf{OACIRR} dataset.}
    \label{fig:benchmark_construction_pipeline}
    \vspace{-3mm}
\end{figure*}

\section{Related Work}
\label{sec:related_work}

\textbf{Composed Image Retrieval.}
Prevailing supervised Composed Image Retrieval (CIR) methods typically leverage Vision-Language Pre-training (VLP) models for foundational encoding, subsequently employing various adaptation strategies tailored to the retrieval task~\cite{artemis, data_roaming, reranking, sprc, cala, detailfusion}.
To alleviate reliance on annotated triplets, Zero-Shot CIR (ZS-CIR) approaches explore either converting the reference image into a pseudo-text representation~\cite{pic2word, searle, clip_inversion, palavra} or using LLM-generated target descriptions~\cite{cirevl, ldre} to recast the problem as text-to-image retrieval.
Another research line addresses data scarcity by automatically synthesizing large-scale training triplets~\cite{data_roaming, covr, compodiff, ipr2pr, magiclens}.
Despite their differences, these approaches operate at the semantic level and therefore struggle to reliably retrieve a user-specified instance across contexts.
In contrast, our OACIR task imposes strict constraints on instance fidelity, enabling more precise and reliable retrieval.

\noindent
\textbf{Instance Consistency in Image Retrieval.}
Instance-level consistency has long been a central goal in image retrieval, explored extensively within person-centric tasks such as Image-based Person Retrieval (IPR)~\cite{reid_1, reid_2, reid_3, reid_4}, its clothes-changing variants (CC-IPR)~\cite{ccreid}, and more recently, Composed Person Retrieval (CPR)~\cite{cpr}.
While these methods have advanced person identification under various conditions, their specialized focus inherently limits their applicability to broader object categories in general-purpose retrieval.
A distinct paradigm achieves instance awareness by fine-tuning a model to associate a visual concept with a learnable textual token~\cite{palavra, meta, myvlm}.
However, this reliance on per-instance optimization hinders both scalability and practical utility.
In contrast, our OACIR framework achieves robust instance fidelity through an explicit visual prompt at inference time, offering a more flexible and general-purpose approach that bypasses the need for either domain-specific architectures or per-instance fine-tuning.

\section{The OACIRR Benchmark}
\label{sec:the_oacirr_benchmark}

Advancing OACIR requires a benchmark that moves beyond semantic-level matching to enforce strict instance-level consistency.
To this end, we propose a comprehensive pipeline for constructing OACIR data from real-world images, as detailed in Section~\ref{subsec:dataset_construction}.
Leveraging this pipeline, we construct \textbf{OACIRR} (\textbf{O}bject-\textbf{A}nchored \textbf{C}omposed \textbf{I}mage \textbf{R}etrieval on \textbf{R}eal-world images), a pioneering large-scale, multi-domain benchmark for this emergent task.
A comprehensive analysis of its quality, diversity, and the challenges it poses is presented in Section~\ref{subsec:dataset_analysis}.

\subsection{Dataset Construction}
\label{subsec:dataset_construction}

As illustrated in Figure~\ref{fig:benchmark_construction_pipeline}, our \textbf{OACIRR} dataset construction pipeline comprises four sequential key stages: (\textit{i}) Image Pairs Collection, (\textit{ii}) Image Pairs Filtering, (\textit{iii}) Quadruples Annotation, and (\textit{iv}) Candidate Gallery Construction.
We detail each stage below:

\noindent
\textbf{Stage 1: Image Pair Collection.}
The foundation of OACIR lies in sourcing image pairs that feature an identical instance across different contexts.
We leverage four large-scale, fine-grained visual classification datasets as our primary sources: DeepFashion2~\cite{deepfashion2}, Stanford Cars~\cite{stanford_cars}, Products-10K~\cite{products-10k}, and Google Landmarks v2~\cite{google_landmarks}.
Given a source dataset $\mathcal{D} \! = \! \{(I_i, \, y_i)\}_{i=1}^N$, where $I_i$ represents an image and $y_i$ denotes its instance-level ID, we first organize the images with the same ID into high-fidelity sets $\mathcal{S}_j = \{I_i | y_i = y_j\}$ by applying fine-grained classification and visual consistency filtering.
Subsequently, a set $\mathcal{S}_j$ is considered valid for construction if it contains at least $\tau_{valid}$ images.
All construction-valid image sets proceed to the subsequent quadruple construction stages, while the remainder are reserved for populating the candidate gallery.

\begin{table*}[t]
    \vspace{-3.5mm}
    \centering
    \setlength{\tabcolsep}{5pt}
    \renewcommand{\arraystretch}{1.0}
    \resizebox{\textwidth}{!}{
        \begin{tabular}{lcccccccccc}
        \toprule
        \multirow{2}{*}{ \textbf{Dataset} } & \multirow{2}{*}{ \textbf{Publication} } & \multirow{2}{*}{ \textbf{\# Samples} } & \multirow{2}{*}{ \textbf{Splits} } & \multirow{2}{*}{ \textbf{Data Type} } & \textbf{Avg Length of} & \textbf{Instance} & \textbf{Instance} & \textbf{Visual} & \textbf{Contextual} & \textbf{Multi-} \\
          &  &  &  &  & \textbf{Modification Text} & \textbf{Consistency} & \textbf{Distractors} & \textbf{Grounding} & \textbf{Modification Text} & \textbf{Domain} \\
        \midrule
        \rowcolor{gray!10}
        CIRR~\cite{cirr} & ICCV 2021 & 36.6K & train, eval & real-world & 11.3 & \textcolor{markred}{\XSolidBrush} & \textcolor{markred}{\XSolidBrush} & \textcolor{markred}{\XSolidBrush} & \textcolor{markred}{\XSolidBrush} & \textcolor{teal}{\Checkmark} \\

        \rowcolor{gray!10}
        FashionIQ~\cite{fashioniq} & CVPR 2021 & 30.1K & train, eval & real-world & 5.3 & \textcolor{markred}{\XSolidBrush} & \textcolor{teal}{\Checkmark} & \textcolor{markred}{\XSolidBrush} & \textcolor{markred}{\XSolidBrush} & \textcolor{markred}{\XSolidBrush} \\

        \rowcolor{gray!10}
        CIRCO~\cite{searle} & ICCV 2023 & 1.0K & eval & real-world & 8.2 & \textcolor{markred}{\XSolidBrush} & \textcolor{markred}{\XSolidBrush} & \textcolor{markred}{\XSolidBrush} & \textcolor{markred}{\XSolidBrush} & \textcolor{teal}{\Checkmark} \\

        InstructPix2Pix~\cite{ipr2pr} & CVPR 2023 & 454K & train, eval & synthetic & 9.4 & \textcolor{markred}{\XSolidBrush} & \textcolor{markred}{\XSolidBrush} & \textcolor{markred}{\XSolidBrush} & \textcolor{markred}{\XSolidBrush} & \textcolor{teal}{\Checkmark} \\

        LaSCo~\cite{data_roaming} & AAAI 2024 & 389K & train & synthetic & 5.9 & \textcolor{markred}{\XSolidBrush} & \textbf{--} & \textcolor{markred}{\XSolidBrush} & \textcolor{markred}{\XSolidBrush} & \textcolor{teal}{\Checkmark} \\

        CIRHS~\cite{cirhs} & ACM MM 2025 & 535K & train & synthetic & 10.2 & \textcolor{markred}{\XSolidBrush} & \textbf{--} & \textcolor{markred}{\XSolidBrush} & \textcolor{teal}{\Checkmark} & \textcolor{teal}{\Checkmark} \\

        SynCPR~\cite{cpr} & NIPS 2025 & 1.1M & train & synthetic & 13.3 & \textcolor{teal}{\Checkmark} & \textbf{--} & \textcolor{markred}{\XSolidBrush} & \textcolor{teal}{\Checkmark} & \textcolor{markred}{\XSolidBrush} \\

        ITCPR~\cite{cpr} & NIPS 2025 & 2.2K & eval & real-world & 9.5 & \textcolor{teal}{\Checkmark} & \textcolor{teal}{\Checkmark} & \textcolor{markred}{\XSolidBrush} & \textcolor{markred}{\XSolidBrush} & \textcolor{markred}{\XSolidBrush} \\

        \rowcolor{orange!10} \textbf{OACIRR (Ours)} & \textbf{CVPR 2026} & \textbf{161K} & \textbf{train, eval} & \textbf{real-world} & \textbf{20.1} & \textcolor{teal}{\Checkmark} & \textcolor{teal}{\Checkmark} & \textcolor{teal}{\Checkmark} & \textcolor{teal}{\Checkmark} & \textcolor{teal}{\Checkmark} \\
        \bottomrule
        \end{tabular}
    }

    \vspace{-0.5mm}
    \caption{Comparative analysis of existing Multimodal Image Retrieval datasets.}
    \label{tab:dataset_comparison}
    \vspace{-3.5mm}

\end{table*}

\noindent
\textbf{Stage 2: Image Pair Filtering.}
To ensure quadruple quality and task difficulty, we perform a rigorous two-step filtering process on the image pairs sampled from each set $\mathcal{S}_j$.
First, to ensure the modification text is meaningful and to prevent models from relying on trivial image similarity shortcuts, we discard overly similar pairs by thresholding their feature cosine similarity.
Second, to foster richer background diversity, we filter out class-centric images.
Specifically, an image is discarded if it is visually similar to at least $\tau_{count}$ other images within the same set.

\begin{figure}[t]
    \vspace{-0.25mm}
    \centering
        \includegraphics[width=0.98\linewidth, trim=0cm 0cm 0cm 0cm, clip]{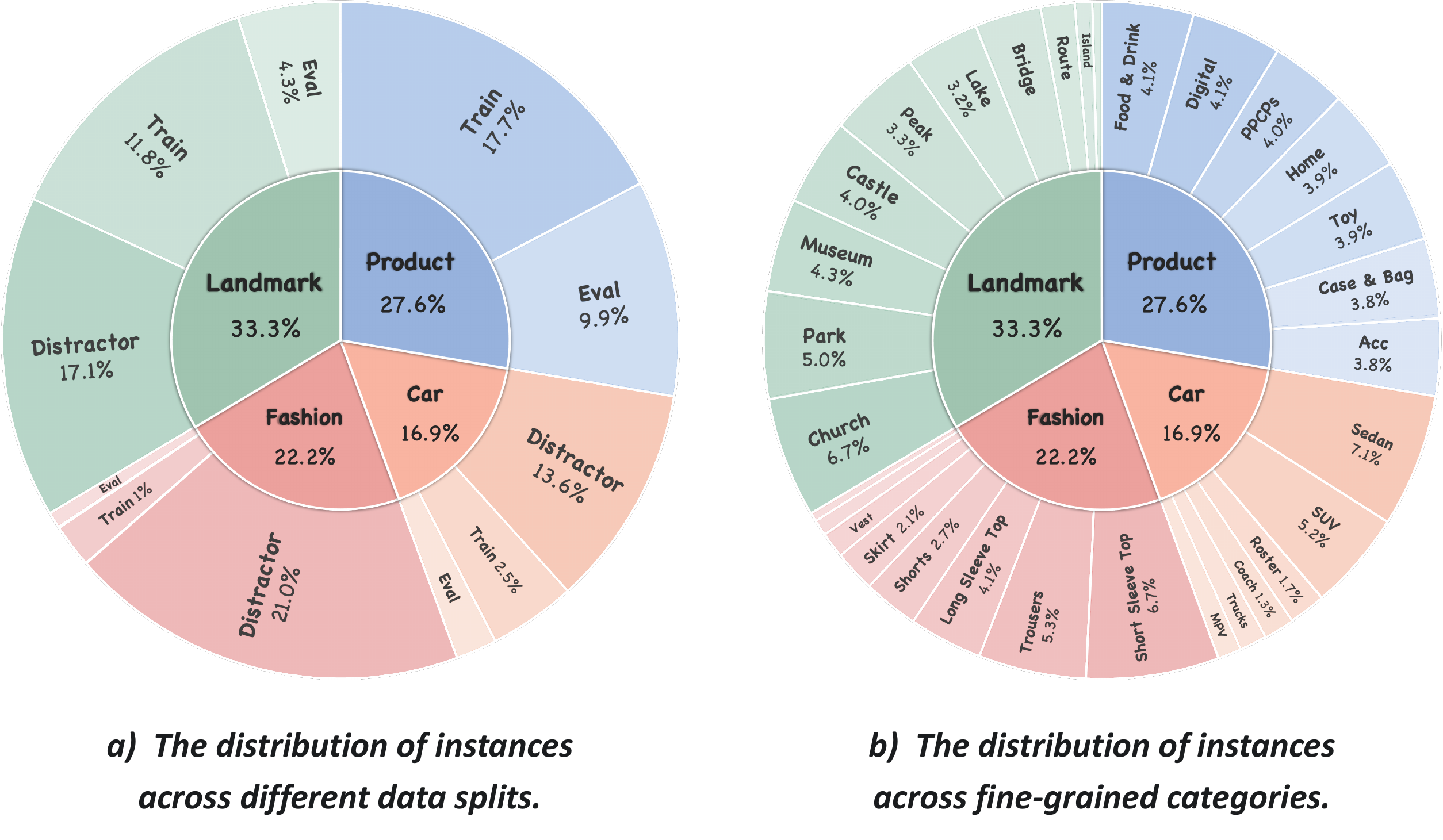}
    \vspace{-2.75mm}
    \caption{Instance distribution of the \textbf{OACIRR} benchmark.}
    \label{fig:instance_distribution}
    \vspace{-3.75mm}
\end{figure}

\noindent
\textbf{Stage 3: Quadruple Annotation.}
From each filtered pair of the reference and target image $(I_r, I_t)$, we conduct a semi-automatic annotation process to construct the final quadruple $(I_r, B_r, T_m, I_t)$, where $B_r$ denotes the bounding box of the anchored instance on $I_r$, and $T_m$ is the modification text.
We first leverage a powerful MLLM~\cite{qwen-vl-max} to generate both the modification text $T_m$ and the instance's class label $l_{ins}$.
For bounding box annotation, we employ a grounding model~\cite{mm_grounding_dino} to generate initial proposals.
Proposals with confidence scores below a predefined threshold are then manually annotated to ensure ground-truth precision.
Finally, the entire corpus of annotated quadruples is partitioned into training and evaluation sets at an 8:2 ratio.

\noindent
\textbf{Stage 4: Candidate Gallery Construction.}
To rigorously evaluate a model's instance discrimination capabilities, we construct a dedicated candidate gallery $\mathcal{G}_s$ for each of the four subsets $s$ in the evaluation benchmark.
Each gallery comprises the complete set of ground-truth target images $\{I_t\}$ from the test quadruples of subset $s$, supplemented by a curated collection of distractors.
To maximize instance-level ambiguity, distractors are sourced via a targeted hard-negative mining strategy:
We first identify the set of all unique category labels $\mathcal{L}_s$ present within the test queries of the subset $s$.
We then populate the gallery with hard negatives by sampling images from the reserved pool (from Stage 1) with category labels $l_{cat} \in \mathcal{L}_s$.
This strategy ensures that each gallery is densely enriched with distractors that are categorically relevant but instance-inconsistent.

\begin{figure*}[t]
    \vspace{-1mm}
    \centering
        \includegraphics[width=0.98\linewidth]{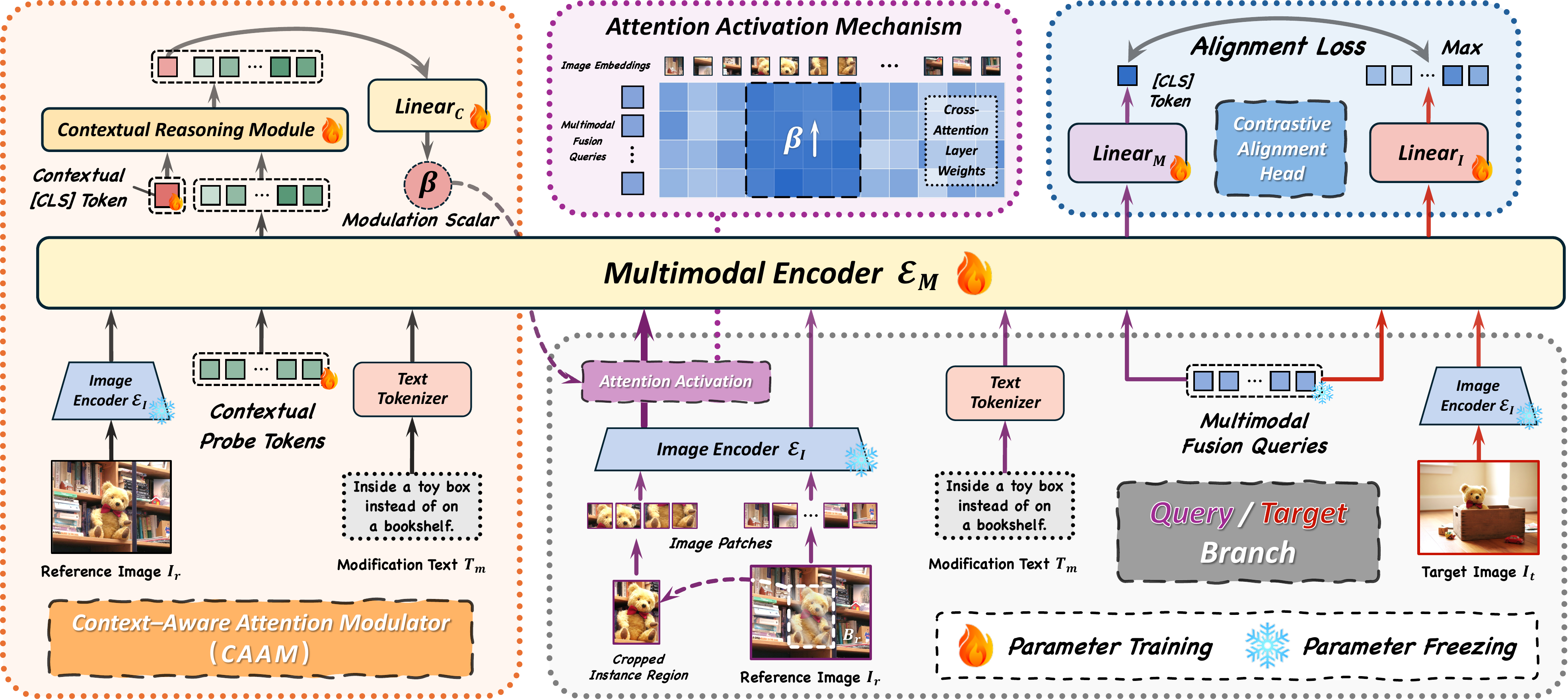}
    \caption{Overall architecture of our proposed \textbf{\textit{AdaFocal}} framework.}
    \label{fig:adafocal_framework}
    \vspace{-2mm}
\end{figure*}

\subsection{Dataset Analysis}
\label{subsec:dataset_analysis}

We provide a comprehensive analysis of the \textbf{OACIRR} benchmark from three perspectives: (\textit{i}) Quality and Contributions, (\textit{ii}) Diversity and Statistics, (\textit{iii}) Core Challenges.

\noindent
\textbf{Quality and Contributions.}
As summarized in Table~\ref{tab:dataset_comparison}, \textbf{OACIRR} establishes a new standard for identity-preserving compositional retrieval through several pivotal features.
(1) \textit{Real-World Authenticity}: Sourced entirely from real-world images, it sets a new benchmark for authentic scenes that directly reflect practical application scenarios.
(2) \textit{Instance-level Fidelity}: The benchmark is built upon the principle of \textit{Instance Consistency}, ensuring every quadruple maintains the anchored instance's precise identity.
This principle is reinforced by a candidate gallery enriched with targeted \textit{Instance Distractors}, creating a challenging testbed for fine-grained discrimination.
(3) \textit{Enhanced Usability}: \textbf{OACIRR} pioneers the integration of \textit{Visual Grounding} via bounding boxes, providing an explicit, non-verbal cue that enhances both query precision and user convenience.
(4) \textit{Modality Synergy}: The dense modification texts, which describe contextual changes, foster a strong synergistic interplay between the visual and textual modalities, compelling models to perform genuine compositional reasoning.

\noindent
\textbf{Diversity and Statistics.}
\textbf{OACIRR} provides a complete ecosystem for model development, featuring a large-scale training set of over 127K quadruples from 2,647 unique instances, and a multi-domain evaluation benchmark with 33.4K quadruples across 1,238 unique instances.
As illustrated in Figure~\ref{fig:instance_distribution}, the instances are distributed across four distinct domains, a curated design intended to evaluate both retrieval depth and breadth.
The Fashion, Car, and Landmark subsets evaluate \textit{retrieval depth}, featuring densely curated galleries of approximately 5K candidates each, drawn from over 1,000 distractor IDs to challenge a model's ability to discriminate between highly similar instances.
In contrast, the Product subset tests \textit{retrieval breadth}, with a vast gallery of nearly 12K candidates from 800 unique IDs that assesses a model's efficiency and accuracy at scale.

\noindent
\textbf{Core Challenges.}
Successfully addressing the \textbf{OACIRR} benchmark demands a sophisticated set of capabilities from the retrieval models.
Specifically, models must demonstrate:
(1) \textit{Advanced Compositional Reasoning}: The ability to perceive subtle visual details and comprehend complex modification texts, and to fuse them into a unified representation.
(2) \textit{Fine-grained Instance Discrimination}: The ability to distinguish a specific visual instance from a gallery saturated with semantically and visually similar distractors.
(3) \textit{Adaptive Visual Attention}: The ability to interpret the bounding box as a visual prompt and dynamically intensify focus within the region while preserving the compositional context.
Collectively, these challenges establish \textbf{OACIRR} as a rigorous benchmark for advancing the frontier of identity-preserving compositional retrieval.

\section{Method}
\label{sec:method}

To address the core challenges of the OACIR task, we propose \textbf{\textit{AdaFocal}}, an effective framework that dynamically modulates visual attention for precise, instance-level retrieval.
Our approach augments a multimodal fusion backbone with a dedicated module that learns to adaptively focus on user-specified instance regions, enabling a nuanced balance between instance fidelity and compositional reasoning.

\subsection{Overall Architecture}
\label{subsec:overall_architecture}

As illustrated in Figure~\ref{fig:adafocal_framework}, \textbf{\textit{AdaFocal}} is built around a central Multimodal Encoder $\mathcal{E_M}$, which serves as the backbone for both query and target feature extraction.

The framework's design reflects a two-stage reasoning process:
(1) \textbf{\textit{Contextual Perception}}: It first \textit{perceives} and \textit{reasons} over the query's compositional context via the \textit{Context-Aware Attention Modulator (CAAM)}.
(2) \textbf{\textit{Adaptive Focus}}: It then \textit{dynamically focuses} on the anchored instance to generate the final composed representation for retrieval.

The framework operates through two parallel branches:
\begin{itemize}
\item
    The \textbf{\textit{Query Branch}} processes the input query $(I_r, \! B_r, \! T_m ) \!$. It is uniquely augmented by the \textit{CAAM}, which analyzes the multimodal context to predict a modulation signal. This signal drives the \textit{Attention Activation Mechanism}, which amplifies the focus on the specified instance region during feature fusion within the multimodal encoder $\mathcal{E_M}$.
\item
    The \textbf{\textit{Target Branch}} processes the target image $I_t$ through the same frozen Image Encoder $\mathcal{E_I}$ and multimodal encoder $\mathcal{E_M}$ to produce its representation.
\end{itemize}

Finally, the output representations from both branches are projected into a shared embedding space by a \textit{Contrastive Alignment Head} for similarity computation.

\begin{table*}[t]
    \centering
    \setlength{\tabcolsep}{4pt}
    \renewcommand{\arraystretch}{1.2}
    \resizebox{\textwidth}{!}{
        \begin{tabular}{c|c|c|ccc|ccc|ccc|ccc|c}
        \toprule

        \multirow{2.4}{*}{\large\textbf{Domain}} & \multirow{2.4}{*}{\large\textbf{Method}} & \multirow{2.4}{*}{\textbf{Pretraining Data}} & \multicolumn{3}{c|}{\large\textcolor{deepfashion}{\textbf{Fashion}}} & \multicolumn{3}{c|}{\large\textcolor{cars}{\textbf{Car}}} & \multicolumn{3}{c|}{\large\textcolor{products}{\textbf{Product}}} & \multicolumn{3}{c|}{\large\textcolor{landmarks}{\textbf{Landmark}}} & \multirow{2.4}{*}{\large\textbf{\textit{Avg.}}} \\

        \cmidrule(lr){4-6} \cmidrule(lr){7-9} \cmidrule(lr){10-12} \cmidrule(lr){13-15}

        &  &  & \textbf{$\text{R}_{\text{ID}}\!\text{@1}$} & \textbf{R@1} & \textbf{R@5} & \textbf{$\text{R}_{\text{ID}}\!\text{@1}$} & \textbf{R@1} & \textbf{R@5} & \textbf{$\text{R}_{\text{ID}}\!\text{@1}$} & \textbf{R@1} & \textbf{R@5} & \textbf{$\text{R}_{\text{ID}}\!\text{@1}$} & \textbf{R@1} & \textbf{R@5} & \\

        \midrule

        \multirow{7}{*}{\large\textbf{UMR}}
        & $\text{UniIR-CLIP}_{\textit{SF}}$~\cite{uniir} & \multirow{2}{*}{M-BEIR~\cite{uniir}} & 17.33 & 12.26 & 24.76 & 32.67 & 16.95 & 41.89 & 33.71 & 18.22 & 40.10 & 29.47 & 15.51 & 43.24 & 27.18 \\
        & $\text{UniIR-BLIP}_{\textit{FF}}$~\cite{uniir} &  & 28.53 & 22.41 & 39.63 & 37.21 & 19.97 & 46.51 & 37.76 & 20.98 & 43.19 & 31.71 & 17.14 & 52.12 & 33.10 \\
        & LamRA-Ret~\cite{lamra} & M-BEIR + NLI~\cite{nli} & 27.45 & 21.63 & 37.10 & 61.03 & 35.44 & 74.51 & 69.45 & 39.53 & 70.25 & 58.64 & 32.58 & 68.74 & 49.70 \\
        & MM-Embed~\cite{mm_embed} & M-BEIR + MTEB~\cite{mteb} & 41.38 & 34.55 & 52.50 & 53.21 & 30.06 & 62.80 & 71.03 & 41.47 & 71.15 & 78.85 & 38.88 & 79.32 & 54.60 \\
        & GME (\,2B\,)~\cite{gme} &\multirow{2}{*}{UMRB~\cite{gme}} & 38.13 & 32.14 & 51.50 & 58.84 & 31.60 & 66.03 & 76.89 & 44.11 & 74.20 & 73.86 & 38.99 & 75.61 & 55.16 \\
        & GME (\,7B\,)~\cite{gme} &  & 44.98 & 39.24 & 60.18 & 63.11 & 38.34 & 75.38 & 83.44 & 54.60 & 84.15 & 77.11 & 47.09 & 82.69 & 62.53 \\
        & U-MARVEL~\cite{u_marvel} & M-BEIR + NLI & 46.05 & 40.38 & 60.59 & 62.92 & 39.96 & 74.90 & 83.26 & 54.69 & 84.13 & 69.81 & 37.67 & 73.08 & 60.62 \\

        \midrule

        \multirow{2}{*}{\large\textbf{ZS-CIR}}
        & Pic2Word~\cite{pic2word} & \multirow{2}{*}{CC3M~\cite{cc3m}} & 14.98 & 11.15 & 21.55 & 12.07 & 4.07 & 11.32 & 45.95 & 13.66 & 34.19 & 55.98 & 20.99 & 52.12 & 24.84 \\
        & LinCIR~\cite{lincir} &  & 15.78 & 12.04 & 21.82 & 5.55 & 2.23 & 7.28 & 47.55 & 14.63 & 34.91 & 42.76 & 19.57 & 47.15 & 22.61 \\

        \midrule

        \multirow{4}{*}{\large\textbf{CIR}}
        & \multirow{2}{*}{SPRC (ViT-L)~\cite{sprc}} & CIRR~\cite{cirr} & 28.54 & 25.49 & 44.26 & 22.47 & 15.23 & 36.78 & 52.55 & 33.35 & 61.47 & 37.31 & 24.20 & 49.99 & 35.97 \\
        &  & \textbf{OACIRR (Ours)} & 61.09 & 54.80 & 75.85 & 68.99 & 46.48 & 86.95 & 80.29 & 67.14 & 90.41 & 72.62 & 54.27 & 86.11 & 70.42 \\
        \cmidrule(lr){2-16}
        & \multirow{2}{*}{SPRC (ViT-G)~\cite{sprc}} & CIRR~\cite{cirr} & 28.62 & 25.79 & 44.48 & 25.13 & 15.92 & 37.06 & 54.39 & 34.85 & 62.31 & 40.41 & 26.29 & 52.39 & 37.30 \\
        &  & \textbf{OACIRR (Ours)} & 65.25 & 58.51 & 80.89 & 72.87 & 49.82 & 89.57 & 86.05 & 70.61 & 93.68 & 76.32 & 56.04 & 89.00 & 74.05 \\

        \midrule

        \rowcolor{gray!10}
        & \textbf{\textit{AdaFocal} (ViT-L)} &  & 72.60 & 61.95 & 85.30 & 75.68 & 51.87 & 90.04 & 87.76 & 69.94 & 93.32 & 80.50 & 57.55 & 90.25 & 76.40 \\
        \rowcolor{orange!10}
        \multirow{-2}{*}{\large\textbf{OACIR}}
        & \textbf{\textit{AdaFocal} (ViT-G)} & \multirow{-2}{*}{\textbf{OACIRR (Ours)}} & \textbf{77.15} & \textbf{65.31} & \textbf{86.88} & \textbf{78.42} & \textbf{53.63} & \textbf{92.22} & \textbf{91.86} & \textbf{74.11} & \textbf{95.39} & \textbf{82.92} & \textbf{58.47} & \textbf{91.63} & \textbf{79.00} \\

        \bottomrule
        \end{tabular}
    }

    \caption{\textbf{Quantitative comparison on the OACIRR benchmark.} ``Avg.'' represents the average results across all evaluation metrics.}
    \label{tab:oacirr_benchmark}
    \vspace{-2mm}
\end{table*}

\subsection{Context-Aware Attention Modulator}
\label{subsec:context-aware_attention_modulator}

The core challenge in OACIR is to determine the appropriate degree of focus on the instance specified by $B_r$, which should vary based on the semantic context of $I_r$ and $T_m$.
The \textit{CAAM} is designed to address this by making the attention modulation process context-aware and learnable.

As illustrated in the left part of Figure~\ref{fig:adafocal_framework}, the \textit{CAAM} first processes the reference image and modification text via the frozen Image Encoder and a Text Tokenizer.
These features are then fed into the shared multimodal encoder alongside a set of $K$ learnable \textit{Contextual Probe Tokens} (denoted as $\{\text{p}_k\}_{k=1}^K$), which learn contextual cues by interacting with the multimodal inputs.
The resulting output features, along with a learnable \textit{Contextual [CLS] Token}, are then processed by the \textit{Contextual Reasoning Module (CRM)}.
The \textit{CRM} aggregates and reasons over these tokens to produce a final contextual representation, which is then projected by a mapping layer, $\text{Linear}_\mathcal{C}(\cdot)$, to form the final query-specific \textit{Modulation Scalar}~$\beta$ for adaptive attention modulation:
\begin{equation}
    \beta = \text{Linear}_\mathcal{C}(\text{CRM}(\mathcal{E_M}(\mathcal{E_I}(I_r), T_m, \{\text{p}_k\}))).
\label{eq:1}
\end{equation}

\subsection{Attention Activation Mechanism}
\label{subsec:attention_activation_mechanism}

The modulation scalar generated by the \textit{CAAM} drives the \textit{Attention Activation Mechanism} within the query branch.

The multimodal encoder fuses visual information via cross-attention between its $M$ frozen Multimodal Fusion Queries, denoted as $\{\text{q}_m\}_{m=1}^M$, and the $N$ visual patch embeddings $\{\text{e}_n\}_{n=1}^N$ from the reference image.

Inspired by attention manipulation techniques developed for generative models~\cite{prompt_highlighter}, we adapt this principle to the retrieval task by injecting the learned modulation scalar as a dynamic bias into the cross-attention computation.
A binary mask $M_{B_r}$, spatially aligned with the patch embeddings corresponding to the bounding box $B_r$, is used to apply this bias.
The output of the modulated cross-attention, which produces the updated queries $\{\hat{\text{q}}_m\}$, is formulated as:
\begin{equation}
    \{\hat{\text{q}}_m\} = A'V = \text{Softmax} \left( \frac { Q K^T + \beta \cdot M_{B_r} } {\sqrt{d_k}} \right) V,
\label{eq:2}
\end{equation}
where $A'$ denotes the modulated attention weights, $Q = f_q(\{\text{q}_m\})$, $K = f_k(\{\text{e}_n\})$, and $V = f_v(\{\text{e}_n\})$ represent the transformed query, key, and value matrices obtained via projections $f_{(\cdot)}$, and $d_k$ denotes the dimension of matrix $K$.

This adaptive mechanism, driven by the context-aware modulation scalar $\beta$, intensifies the model's focus on the user-specified instance by re-weighting the value matrix $V$.

\subsection{Objective Function}
\label{subsec:objective_function}

During training, the final query representation $f_q$ is obtained by projecting the [CLS] token from the query branch through the multimodal mapping layer $\text{Linear}_\mathcal{M}(\cdot)$:
\begin{equation}
    f_q = \text{Linear}_\mathcal{M}(\mathcal{E'_M}(\mathcal{E_I}(I_r), B_r, T_m, \{q_m\})),
\label{eq:3}
\end{equation}
where $\mathcal{E'_M}$ denotes the multimodal encoder operating with the \textit{Attention Activation Mechanism}.

Similarly, the target representation $f_t$ is obtained by projecting the image tokens from the target branch through the image mapping layer $\text{Linear}_\mathcal{I}(\cdot)$:
\begin{equation}
    f_t = \text{Linear}_\mathcal{I}(\mathcal{E_M}(\mathcal{E_I}(I_t), \{q_m\})).
\label{eq:4}
\end{equation}

The entire framework is trained end-to-end using a batch-based contrastive learning objective.
We employ the Contrastive Alignment Loss, formulated as:
\begin{equation}
    \mathcal{L}_{\text{Align}} = -\frac{1}{|\mathcal{B}|} \sum_{i=1}^{|\mathcal{B}|} \log
    \frac{\mathbb{S}( f_q^{(i)}, f_t^{(i)} )}
    { \sum_{j=1}^{|\mathcal{B}|} \mathbb{S}( f_q^{(i)}, f_t^{(j)} ) },
\label{eq:5}
\end{equation}
where { $\mathbb{S}(a, b) \! \coloneqq \! \exp(Sim( (a, b) / \tau)$ }, $Sim(\cdot)$ represents the cosine similarity between features, ${\tau}$ is a temperature hyper-parameter.
$\mathcal{B}$ denotes the training batch, ${f_q}^{(i)}$ and ${f_t}^{(i)}$ denote the $i$-th query and target representations in $\mathcal{B}$.

During inference, the \textit{CAAM} dynamically predicts the modulation scalar $\beta$ for each unique query, and the resulting query representation $f_q$ is used to rank all candidates in the gallery based on cosine similarity.

\section{Experiments}
\label{sec:experiments}

\subsection{Experimental Setup}
\label{subsec:experimental_setup}

\textbf{Benchmark Details.}
The experiments are conducted primarily on the newly proposed \textbf{OACIRR} benchmark.
The evaluation benchmark comprises four distinct subsets (\textcolor{deepfashion}{Fashion}, \textcolor{cars}{Car}, \textcolor{products}{Product}, and \textcolor{landmarks}{Landmark}), each with dedicated queries and a candidate gallery, totaling 33.4K queries and 26.6K unique images.
The training set is a unified collection of 127.2K quadruples, aggregated from all four subsets, used for fine-tuning models.

\noindent
\textbf{Implementation Details.}
All experiments are conducted on four Tesla V100 GPUs with 32GB of memory.
During \textbf{OACIRR} construction, modification texts were generated using Qwen-VL-Max~\cite{qwen-vl-max}, while bounding boxes were annotated using MM-Grounding-DINO-Large~\cite{mm_grounding_dino}.
For fine-tuning \textbf{\textit{AdaFocal}} on \textbf{OACIRR}, we set the number of epochs to 20 and the batch size to 128.
We employ the AdamW optimizer~\cite{adamw} with betas set to (0.9, 0.98) and a weight decay of 0.05.
The Multimodal Encoder is based on the BLIP-2 Q-Former~\cite{blip2}.
To ensure stable training and balanced parameter updates, a differential learning rate strategy is employed.
The learning rate for the lightweight \textit{CAAM} is set to 1e-4, while the parameters of the Multimodal Encoder are fine-tuned with a smaller learning rate of 1e-5.
The temperature hyper-parameter $\tau$ is set to 0.07.

\noindent
\textbf{Evaluation Metrics.}
The evaluation of OACIR centers on two key aspects: (1) the \textit{semantic correctness} of the final retrieved image and (2) the \textit{consistency} of the anchored instance.
To this end, we introduce the top-K \textit{Instance Recall} ($\text{R}_{\text{ID}}\text{@K}$) in addition to the standard top-K Recall (R@K).
A retrieval is deemed correct under $\text{R}_{\text{ID}}$ only if the retrieved image contains the exact same instance specified within the reference image's bounding box.
We report Recall@1 and Recall@5 to assess overall retrieval performance, alongside $\text{R}_{\text{ID}}\text{@1}$ to specifically measure the model's instance fidelity.

\begin{table}[t]
    \vspace{-1mm}
    \centering
    \small
    \setlength{\tabcolsep}{7pt}
    \renewcommand{\arraystretch}{1.18}
    \resizebox{\linewidth}{!}{
        \begin{tabular}{cc|cccc}
        \toprule

        \multicolumn{2}{c|}{\textbf{CAAM}} & \multicolumn{4}{c}{\textbf{OACIRR Benchmark}} \\
        \cmidrule(lr){1-2} \cmidrule(lr){3-6}
        \textbf{CRM} & \textbf{Probe Tokens} & $\text{R}_{\text{ID}}\!\text{@1}$ & R@1 & R@5 & \textbf{Avg.} \\

        \midrule

        \rowcolor{gray!10}
        \multicolumn{2}{c|}{ $\textbf{\textit{Baseline}} \,\,\, \textit{(w/o CAAM)}$ } & 77.74 & 58.39 & 88.61 & 74.91 \\

        \midrule

        \multirow{2}{*}{\textbf{Average Pooling}} & \textcolor{frozen}{\textbf{\textit{Frozen}}} & 79.70 & 59.84 & 89.62 & 76.39 \\
        & \textcolor{learnable}{\textbf{\textit{Learnable}}} & 79.83 & 59.57 & 89.54 & 76.31 \\

        \midrule

        \multirow{2}{*}{\textbf{MLP}} & \textcolor{frozen}{\textbf{\textit{Frozen}}} & 80.51 & 60.55 & 90.15 & 77.07 \\
        & \textcolor{learnable}{\textbf{\textit{Learnable}}} & 81.10 & 61.10 & 90.40 & 77.53 \\

        \midrule

        & \textcolor{frozen}{\textbf{\textit{Frozen}}} & 81.59 & 61.85 & 91.13 & 78.19 \\
        \rowcolor{orange!10}
        \multirow{-2}{*}{\textbf{Transformer}} & \textcolor{learnable}{\textbf{\textit{Learnable}}} & \textbf{82.59} & \textbf{62.88} & \textbf{91.53} & \textbf{79.00} \\

        \bottomrule
        \end{tabular}
    }

    \caption{Ablation study on the architecture of the \textit{CAAM}.}
    \label{tab:ablation_caam}
    \vspace{-3mm}
\end{table}

\subsection{Quantitative Evaluation}
\label{subsec:quantitative_evaluation}

We present a comprehensive quantitative evaluation on the \textbf{OACIRR} benchmark to analyze the capabilities of existing retrieval paradigms and demonstrate the effectiveness of our proposed \textbf{\textit{AdaFocal}} framework.
As detailed in Table~\ref{tab:oacirr_benchmark}, we assess three distinct groups: Universal Multimodal Retrieval (UMR) models, Composed Image Retrieval (CIR) methods, and our proposed approach.

\noindent
\textbf{Evaluation Settings.}
To ensure a fair comparison, the evaluation protocol is adapted for each model class:
\textbf{(1)} For \textbf{UMR models}, capable of visual grounding, the reference image is rendered with the bounding box, accompanied by a textual prompt explicitly instructing the model to preserve the anchored instance.
\textbf{(2)} For \textbf{Zero-shot and Supervised CIR methods}, which lack native support for bounding box inputs, the OACIR task is converted into a standard CIR format by embedding the instance's unique ID tag into the modification text.
\textbf{(3)} Our \textbf{\textit{AdaFocal}} framework is inherently designed to process the native OACIR task input.

\begin{figure}[t]
    \vspace{-1mm}
    \centering
        \includegraphics[width=\linewidth, trim=0cm 0.7cm 0cm 0.3cm, clip]{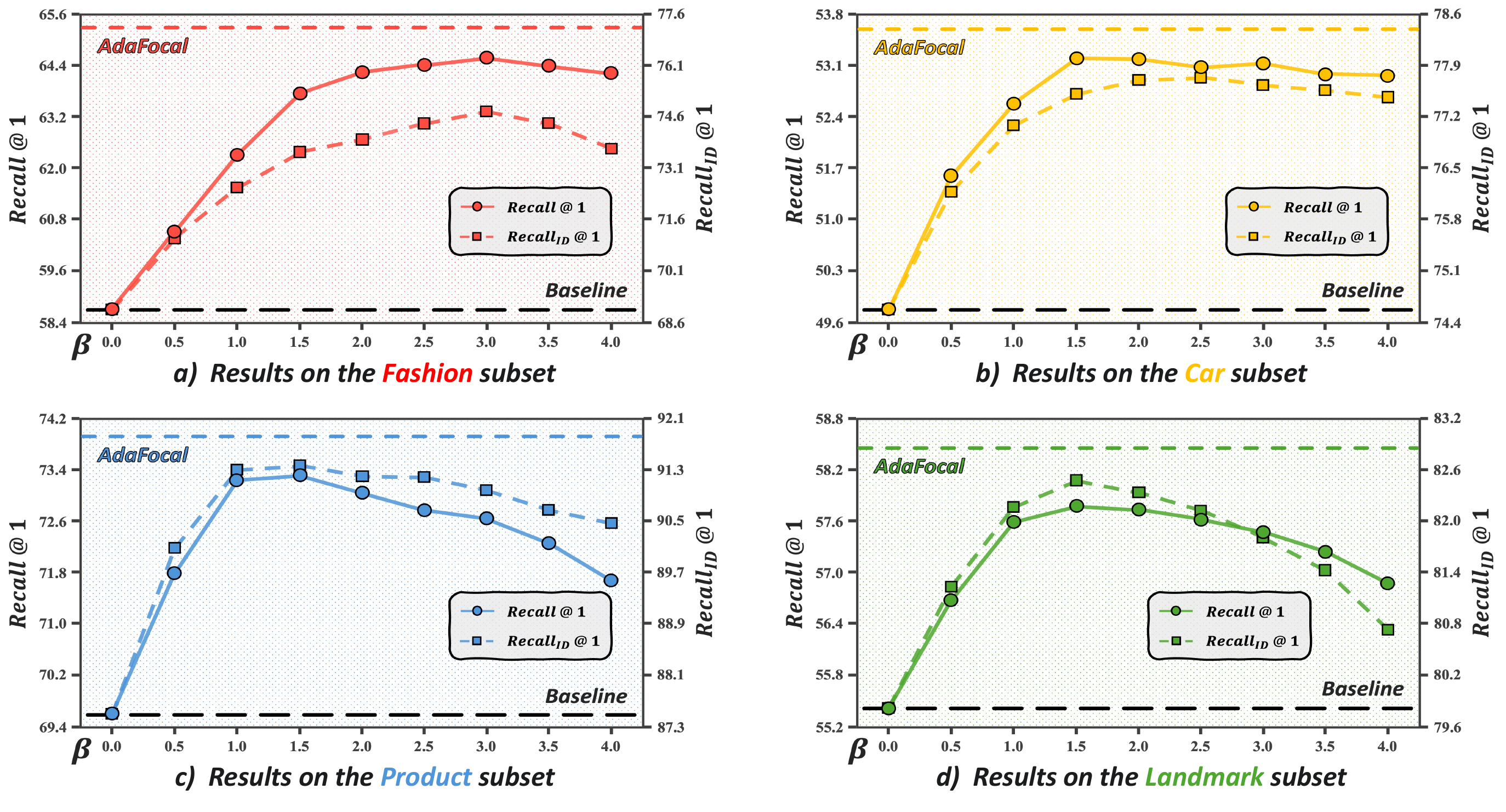}
    \vspace{=-4.5mm}
    \caption{Ablation study on the \textit{Modulation Scalar}~$\beta$.}
    \label{fig:ablation_beta}
    \vspace{-3mm}
\end{figure}

\begin{figure*}[t]
    \centering
        \includegraphics[width=\linewidth, trim=0.6cm 0.25cm 0cm 0.5cm, clip]{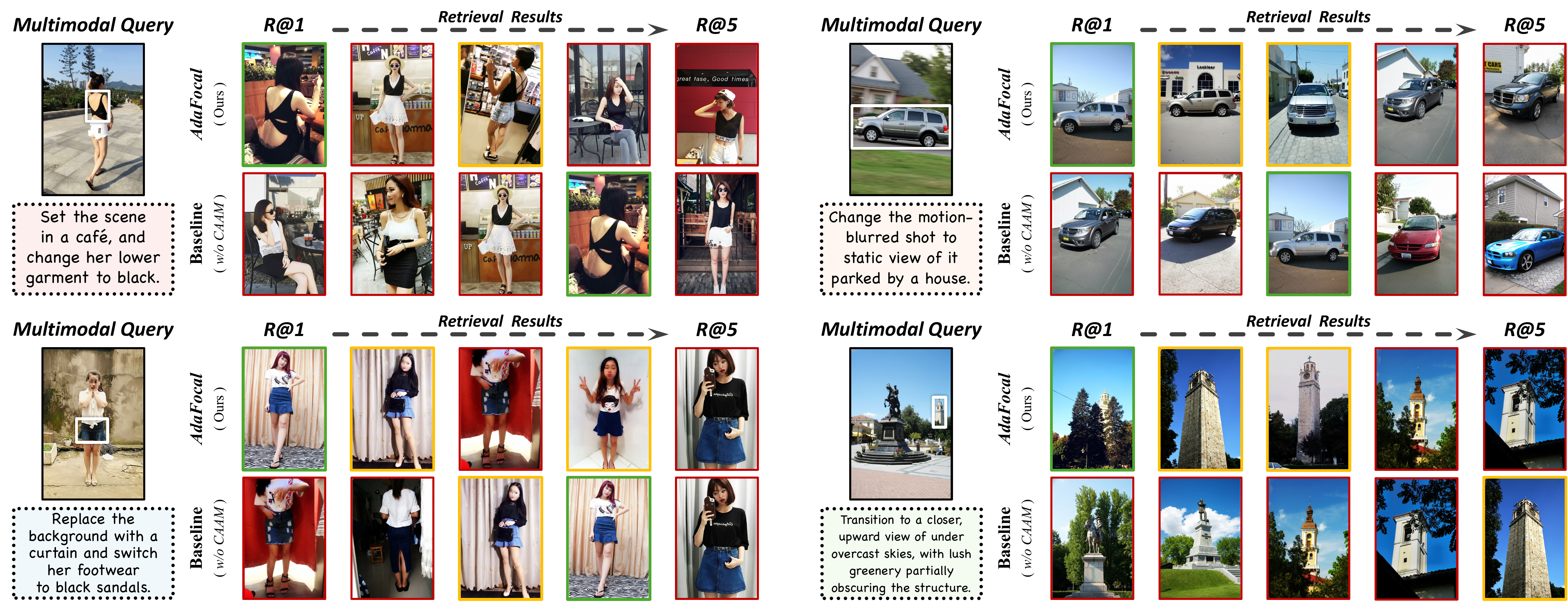}
    \caption{Qualitative comparison of our \textbf{\textit{AdaFocal}} and the Baseline on the \textbf{OACIRR} benchmark. \textbf{\textcolor[HTML]{4EA72E}{Green boxes}} indicate the ground-truth target, \textbf{\textcolor[HTML]{FFC000}{yellow boxes}} indicate instance-correct but semantically incorrect results, and all other retrieved images are marked with \textbf{\textcolor[HTML]{C00000}{red boxes}}.}
    \label{fig:qualitative_comparison}
    \vspace{-2mm}
\end{figure*}

\noindent
\textbf{Analysis of Existing Paradigms.}
The results under zero-shot evaluation reveal the profound challenge OACIR poses to existing models.
Even with explicit visual and textual cues, powerful UMR models exhibit limited instance-level fidelity.
Their pre-training prioritizes broad semantic correspondence across diverse multimodal data and therefore does not equip them with the robust instance-level discrimination required for this task, a deficiency particularly evident in multi-object scenarios such as the Fashion subset.
ZS-CIR methods, relying solely on semantic-level textual cues, perform even worse, as they lack the fine-grained visual input necessary to resolve the instance-level ambiguity presented by our benchmark's hard-negative distractors.

\noindent
\textbf{The Critical Role of Instance-Aware Training.}
To isolate the contribution of our dataset, we fine-tune a strong supervised CIR baseline, SPRC~\cite{sprc}.
When trained on the CIRR dataset, SPRC achieves a modest 37.30\% average recall, confirming that semantic-level compositional training is insufficient for OACIR.
However, when the same model is fine-tuned on our \textbf{OACIRR} dataset, its performance soars to 74.05\% in average recall.
This substantial improvement validates the critical role of our dataset's instance-consistent construction in successfully addressing the OACIR task.

\noindent
\textbf{The Efficacy of \textit{AdaFocal}.}
Building upon the strong foundation of our training data, \textbf{\textit{AdaFocal}} demonstrates a further significant performance leap across all subsets.
With an identical ViT-G backbone, it outperforms the OACIRR-trained SPRC by a large margin, achieving average improvements of \textbf{+4.14} in R@1 and \textbf{+7.47} in $\text{R}_{\text{ID}}\text{@1}$.
This gain confirms that our direct and adaptive visual grounding mechanism is more effective than relying on ambiguous textual prompts for instance preservation.
Critically, the narrow gap between R@1 and $\text{R}_{\text{ID}}\text{@1}$ across all baselines indicates their primary failure mode is instance misidentification.
In contrast, \textbf{\textit{AdaFocal}} widens this gap by achieving much higher Instance Recall, demonstrating a superior capability to precisely identify the target instance.

\subsection{Ablation Study}
\label{subsec:ablation_study}

We now dissect the core mechanisms of our \textbf{\textit{AdaFocal}} framework, analyzing the architectural design of the \textit{CAAM} and the impact of the \textit{Attention Activation Mechanism}.

\noindent
\textbf{Component Analysis of CAAM.}
To validate the architectural design of the \textit{CAAM}, we evaluate several variants, with the results presented in Table~\ref{tab:ablation_caam}.
The analysis reveals two key insights.
First, the method of \textit{contextual aggregation} is crucial.
The superiority of the Transformer-based CRM over simpler aggregation methods underscores the necessity of its reasoning capabilities for interpreting complex compositional contexts and predicting a meaningful modulation scalar.
Second, employing \textit{learnable Contextual Probe Tokens} is vital.
Across all configurations, learnable probe tokens consistently outperform their frozen counterparts, with the performance gain being most pronounced when paired with the Transformer CRM.
This highlights a synergistic effect in which advanced reasoning is required to fully exploit the nuanced cues captured by task-adapted probe tokens.

\noindent
\textbf{Efficacy of Adaptive Attention.}
To demonstrate the efficacy of our adaptive focus strategy, we compare \textbf{\textit{AdaFocal}} against a baseline without attention modulation ($\beta=0$) and against variants using a range of fixed, manually set $\beta$ values.
As illustrated in Figure~\ref{fig:ablation_beta}, the results yield three critical findings.
First, applying any positive attention bias ($\beta > 0$) consistently outperforms the baseline, confirming that \textit{explicitly focusing on the anchored instance} is critical for the OACIR task.
Second, a clear \textit{trade-off between instance fidelity and compositional reasoning} emerges as $\beta$ increases.
$\text{R}_{\text{ID}}\text{@1}$ rises sharply and then saturates, demonstrating that intensified focus greatly enhances instance identification.
However, R@1 exhibits a sharper decline after its peak, as an excessively large $\beta$ causes the model to neglect crucial context from both the image background and modification text, leading to semantic mismatches.
Lastly, the optimal fixed $\beta$ varies across subsets, confirming that the ideal balance is highly context-dependent.
Our \textbf{\textit{AdaFocal}} framework, which leverages the \textit{CAAM} to predict a query-specific $\beta$, consistently outperforms any fixed attention activation strategy and operates near the performance ceiling across all conditions.
This provides direct evidence for the necessity of our context-aware, adaptive modulation approach.

\subsection{Qualitative Results}
\label{subsec:qualitative_results}

Qualitative results in Figure~\ref{fig:qualitative_comparison} visually substantiate \textbf{\textit{AdaFocal}}'s superior ability to balance instance fidelity and compositional reasoning.
The baseline model, lacking an adaptive visual attention modulation mechanism, fails by incorrectly prioritizing semantic cues from the modification text and thus retrieves instance-inconsistent results.
In contrast, guided by the \textit{CAAM}, \textbf{\textit{AdaFocal}} retrieves the ground-truth target by adaptively intensifying its focus on the anchored instance, fulfilling the personalization constraint while precisely interpreting the contextual changes.
Notably, the high ranks of other instance-consistent results further underscore our robust instance-level discrimination capabilities.

\section{Conclusion}
\label{sec:conclusion}

In this work, we propose \textbf{O}bject-\textbf{A}nchored \textbf{C}omposed \textbf{I}mage \textbf{R}etrieval (\textbf{OACIR}), a novel task that pushes compositional retrieval beyond semantic matching to achieve rigorous instance fidelity.
To advance research in this emergent area, we construct \textbf{OACIRR}, the first large-scale, multi-domain benchmark that provides a foundational dataset of over 160K real-world quadruples and candidate galleries enriched with curated instance distractors.
Furthermore, we propose \textbf{\textit{AdaFocal}}, a novel framework that dynamically intensifies attention on the anchored instance specified by the bounding box, thereby balancing instance preservation with compositional reasoning.
Extensive experiments validate the task's challenges for existing models while establishing \textbf{\textit{AdaFocal}} as an effective baseline.
We hope that our work inspires a new generation of compositional retrieval systems with greater flexibility and instance-aware reliability.

% To address the unique input format and inherent challenges of OACIR,

% Collectively, we hope these contributions inspire a new generation of more flexible and instance-aware compositional retrieval systems.
% Collectively, we hope our contributions pave the way for compositional retrieval systems with greater flexibility and instance-aware reliability.
% Collectively, we hope our contributions inspire a new generation of compositional retrieval systems that offer greater flexibility while ensuring precise, instance-aware reliability.

% that are more flexible, precise, and instance-aware.

\section*{Acknowledgements}
\label{sec:acknowledgements}

This research was supported by multiple funding sources, including
the Beijing Natural Science Foundation
(Grants L243015,  % 原老师-1
L223003,  % 胡老师-1
and JQ24022),  % 刘老师
the National Natural Science Foundation of China
(Grants 62192782,  % 胡老师-2
62532015,  % 胡老师-3
and 62302501),  % 张老师
and the Beijing Major Science and Technology Project under Contract (No. Z251100008425008).  % 原老师-2

{
    \small
    \bibliographystyle{ieeenat_fullname}
    \bibliography{OACIR}
}

\clearpage
\setcounter{page}{1}
\maketitlesupplementary

\tcbset{
  defaultboxstyle/.style =
    {
        colback=gray!5,
        colframe=black!65,
        coltitle=white,
        fonttitle=\bfseries\center,
        boxrule=0.5mm,
        arc=2.5mm,
        boxsep=1mm,
        left=2mm,
        right=2mm,
        top=2mm,
        bottom=2mm,
        toptitle=0.5mm,
        bottomtitle=0.5mm,
        breakable,
        enhanced,
    }
}

\tcbset{
  fashionboxstyle/.style =
    {
        colback=deepfashion!5,
        colframe=black!65,
        coltitle=white,
        fonttitle=\bfseries\center,
        boxrule=0.5mm,
        arc=2.5mm,
        boxsep=1mm,
        left=2mm,
        right=2mm,
        top=2mm,
        bottom=2mm,
        toptitle=0.5mm,
        bottomtitle=0.5mm,
        breakable,
        enhanced,
    }
}

\tcbset{
  carboxstyle/.style =
    {
        colback=cars!5.5,
        colframe=black!65,
        coltitle=white,
        fonttitle=\bfseries\center,
        boxrule=0.5mm,
        arc=2.5mm,
        boxsep=1mm,
        left=2mm,
        right=2mm,
        top=2mm,
        bottom=2mm,
        toptitle=0.5mm,
        bottomtitle=0.5mm,
        breakable,
        enhanced,
    }
}

\tcbset{
  productboxstyle/.style =
    {
        colback=products!5,
        colframe=black!65,
        coltitle=white,
        fonttitle=\bfseries\center,
        boxrule=0.5mm,
        arc=2.5mm,
        boxsep=1mm,
        left=2mm,
        right=2mm,
        top=2mm,
        bottom=2mm,
        toptitle=0.5mm,
        bottomtitle=0.5mm,
        breakable,
        enhanced,
    }
}

\tcbset{
  landmarkboxstyle/.style =
    {
        colback=landmarks!5,
        colframe=black!65,
        coltitle=white,
        fonttitle=\bfseries\center,
        boxrule=0.5mm,
        arc=2.5mm,
        boxsep=1mm,
        left=2mm,
        right=2mm,
        top=2mm,
        bottom=2mm,
        toptitle=0.5mm,
        bottomtitle=0.5mm,
        breakable,
        enhanced,
    }
}

\section{More Details on the OACIRR Benchmark}
\label{sec:oacirr_benchmark_details}

In this section, we provide a comprehensive overview of the construction pipeline and detailed statistics of the \textbf{OACIRR} benchmark.
We describe the subset-specific protocols in Section~\ref{subsec:subset_specific_construction_pipeline}, the prompts used for MLLM-based annotation in Section~\ref{subsec:mllm_annotation_prompts}, the detailed dataset statistics in Section~\ref{subsec:detailed_dataset_statistics}, and the instance diversity visualization in Section~\ref{subsec:instance_diversity_visualization}.

\subsection{Subset-Specific Construction Pipeline}
\label{subsec:subset_specific_construction_pipeline}

We construct the four \textbf{OACIRR} subsets — \textcolor{deepfashion}{Fashion}, \textcolor{cars}{Car}, \textcolor{products}{Product}, and \textcolor{landmarks}{Landmark} — using four large-scale, fine-grained visual classification datasets: DeepFashion2~\cite{deepfashion2}, Stanford Cars~\cite{stanford_cars}, Products-10K~\cite{products-10k}, and Google Landmarks v2~\cite{google_landmarks}.
Given that these sources differ substantially in structure and granularity, we design tailored protocols and apply subset-specific filtering thresholds throughout the construction pipeline.
We detail each stage below:

\noindent
\textbf{Stage 1: Image Pair Collection.}
The objective of this stage is to establish high-fidelity, instance-level image sets, with procedures tailored to each data source:
\begin{itemize}
\item
    For \textbf{Products-10K}, the images are already organized at the stock-keeping-unit (SKU) level, which naturally aligns with our instance-level fidelity requirement.
\item
    For \textbf{DeepFashion2} and \textbf{Stanford Cars}, the initial groupings (based on item styles or car models) often contain multiple color variants. To obtain color-consistent instance sets, we further subdivide each group using a pre-trained fine-grained classifier (CLIP-ConvNeXt-Base).
\item
    For \textbf{Google Landmarks v2}, image sets vary between \textit{visually coherent views} of a landmark and \textit{knowledge-based collections} that mix disparate appearances. To enforce strict visual consistency, we prompt an MLLM~\cite{qwen-vl-max} to identify and retain only visually coherent subsets.
\end{itemize}

\noindent
\textbf{Stage 2: Image Pair Filtering.}
As summarized in Table~\ref{tab:filtering_thresholds}, we apply subset-specific thresholds to ensure high-quality image pairs and appropriate task difficulty.
A set $\mathcal{S}_j$ is retained only if its size exceeds the construction-valid threshold $\tau_{valid}$.
Image pairs with feature cosine similarity above $\tau_{high}$ are removed to ensure meaningful modifications.
To promote background diversity, an image is filtered out if its feature similarity exceeds $\tau_{centric}$ with at least $\tau_{count}$ other images in the same set.
\begin{itemize}
\item
    To balance the query volume across domains, we adopt smaller $\tau_{valid}$ values for subsets with fewer initial IDs (\textcolor{deepfashion}{Fashion}, \textcolor{cars}{Car}) and larger values for subsets with abundant initial IDs (\textcolor{products}{Product}, \textcolor{landmarks}{Landmark}).
\item
    To calibrate task difficulty across domains, we adopt more relaxed thresholds ($\tau_{high}$, $\tau_{centric}$, $\tau_{count}$) for subsets involving complex multi-object scenes (\textcolor{deepfashion}{Fashion}, \textcolor{landmarks}{Landmark}), and more rigorous thresholds for subsets centered around a single salient object (\textcolor{cars}{Car}, \textcolor{products}{Product}).
\end{itemize}

\begin{table}[t]
    \centering
    \footnotesize
    \setlength{\tabcolsep}{12pt}
    \renewcommand{\arraystretch}{1.4}
    \resizebox{\linewidth}{!}{
        \begin{tabular}{c|cccc}
        \toprule

        \multirow{2.4}{*}{\normalsize{\textbf{Subset}}} & \multicolumn{4}{c}{\normalsize{\textbf{Filtering Threshold}}} \\
        \cmidrule(lr){2-5}
        & \normalsize{$\tau_{valid}$} & \normalsize{$\tau_{high}$} & \normalsize{$\tau_{centric}$} & \normalsize{$\tau_{count}$} \\

        \midrule

        \rowcolor{deepfashion!8}
        \textcolor{deepfashion}{\small{\textbf{Fashion}}} & 8 & 0.92 & 0.88 & 3 \\

        \rowcolor{cars!10}
        \textcolor{cars}{\small{\textbf{Car}}} & 10 & 0.88 & 0.85 & 2 \\

        \rowcolor{products!8}
        \textcolor{products}{\small{\textbf{Product}}} & 20 & 0.88 & 0.85 & 2 \\

        \rowcolor{landmarks!8}
        \textcolor{landmarks}{\small{\textbf{Landmark}}} & 15 & 0.90 & 0.88 & 3 \\

        \bottomrule
        \end{tabular}
    }

    \caption{Filtering Thresholds for each \textbf{OACIRR} subset.}
    \label{tab:filtering_thresholds}
    \vspace{-2.6mm}

\end{table}

\noindent
\textbf{Stage 3: Quadruple Annotation.}
This stage involves a semi-automatic process.
We assign class labels $l_{ins}$ to each high-fidelity instance set using a tailored prompt.
To reinforce the synergy between the visual and textual modalities, we instruct the MLLM to generate modification texts describing only contextual changes, explicitly excluding any mention of the preserved instance.
For bounding boxes, we directly use the ground-truth annotations in DeepFashion2.
For the remaining three subsets, bounding box proposals with confidence scores below 0.3 from our grounding model~\cite{mm_grounding_dino} are manually re-annotated to ensure precision.

\noindent
\textbf{Stage 4: Candidate Gallery Construction.}
To construct challenging yet efficient candidate galleries, we compute the instance class distribution for each test subset.
Each gallery is populated by sampling hard negatives from the reserved image pool (from Stage 1) to match the class distribution of the query set.
This strategy maximizes instance-level ambiguity while maintaining a compact and computationally efficient gallery for the benchmark.

\subsection{MLLM Annotation Prompts}
\label{subsec:mllm_annotation_prompts}

We employed Qwen-VL-Max~\cite{qwen-vl-max} for all MLLM-based annotation tasks, which comprise two key sub-tasks: (1) \textit{generating class labels} for each high-fidelity instance set, and (2) \textit{producing contextual modification text} conditioned on an image pair and its associated instance class label.

\noindent
\textbf{Instance Class Label Generation.}
This step was applied selectively depending on the characteristics of each subset.
For the \textcolor{deepfashion}{Fashion} subset, we directly adopted the coarse-grained apparel categories defined in DeepFashion2.
For the \textcolor{cars}{Car} subset, all instances were uniformly assigned the label ``\texttt{car}''.
Consequently, MLLM-based labeling was required only for the \textcolor{products}{Product} and \textcolor{landmarks}{Landmark} subsets, which exhibit greater category diversity.

For the \textcolor{products}{Product} subset, which involves only class label annotation, the following prompt template was used:

\begin{tcolorbox}[productboxstyle, title=Class Label Generation for Product subset]
Analyze the provided images to identify the single, identical commercial product present in all of them.
Your task is to output a concise, generic tag for this common object. \\

\textbf{Important Context:} \\
    1. There is exactly one object that is the same product across all images. \\
	2. This object may appear in different states, environments, or from different viewing angles in each image. \\

\textbf{Requirements:} \\
	1. Output only the tag for the common object and nothing else. \\
	2. The tag must be a short, descriptive noun phrase in English.
	   It should be specific enough to be unambiguous but not overly detailed. \\
	3. DO NOT include any brand names. \\
	4. DO NOT describe the object's state, its background, the viewing angle, or any similarities or differences between the images. \\
	5. DO NOT include any introductory phrases like ``The common object is:''.
\end{tcolorbox}

For the \textcolor{landmarks}{Landmark} subset, we designed a prompt that concurrently performs visual consistency filtering and class label annotation. The prompt template is as follows:

\begin{tcolorbox}[landmarkboxstyle, title=Visual Consistency Filter \& \\
                                    Class Label Generation for Landmark subset]
Your task is to analyze a set of images from a single landmark ID and determine if they represent a ``Visual-type'' or a ``Knowledge-type'' landmark, based ONLY on the visual evidence provided. \\
When in doubt, classify as ``Knowledge-type''. Your goal is to approve ``Visual-type'' only when the images unambiguously represent a single, consistent landmark, with verification purely from visual cues. \\

\textbf{Landmark Types Explained:} \\
1. \textit{Visual-type}: The images depict a single, visually consistent, and dominant landmark.
The landmark is the same physical entity across all images, even when viewed from different angles or under varying conditions (e.g., day/night, summer/winter).
\end{tcolorbox}

\begin{tcolorbox}[landmarkboxstyle]
2. \textit{Knowledge-type}: The images are related by a shared theme or geographic context but do not contain one visually consistent landmark. Their connection is conceptual or requires external knowledge to identify.
(e.g., different buildings within a university campus; interior and exterior views of a large museum.) \\

\textbf{Response Format:} \\
Your response MUST be a JSON object and nothing else. Follow this exact format: \\
\{ \\
	``type'': ``visual'' or ``knowledge'', \\
	``label'': ``\,Specific Name of the Landmark'' or null, \\
	``reasoning'': ``A brief explanation for your decision.'' \\
\} \\

\textbf{Important Rules:} \\
1. If you classify as ``knowledge'', set ``label'' to null. \\
2. If you classify as ``visual'', provide the class label of the landmark for the ``label''. \\
3. Do not include any introductory text before or after the JSON object.
\end{tcolorbox}

\noindent
\textbf{Contextual Modification Text Generation.}
To ensure that the generated modification text is accurate, diverse, and effectively complements the visual information, we designed domain-specific prompt templates for all four subsets.
A shared instruction across these prompts was to restrict the MLLM to describe only contextual changes, thereby maximizing its synergy with the visual anchor. The corresponding prompt templates are provided below.

\begin{tcolorbox}[fashionboxstyle, title=Modification Text Generation for Fashion subset]
Based on the two provided images, generate a modification text to transform the first image into the second. \\

\textbf{Requirements:} \\
1. The modification text must be written in fluent and natural English, NOT exceeding 30 words. \\
2. Focus exclusively on the most significant and definite changes. DO NOT describe any identical parts between the two images. \\
3. A specific ``\,Object to Ignore'' is provided below. DO NOT mention this object or any of its attributes in the modification text. \\
4. Avoid any explicit references to the images themselves.
   For example, DO NOT use phrases like ``\,in the first image'' or ``\,in the second picture''. \\
5. Employ diverse expressions. Avoid using repetitive sentence structures or fixed grammatical patterns.
\end{tcolorbox}

\begin{tcolorbox}[fashionboxstyle]
\textbf{Examples:} \\
1. The woman is now wearing a large pink bow and holding a light-up wand. \\
2. The person is wearing a denim skirt, and the background changes to a store with shelves and products. \\
3. The girl changed from wearing patterned pants to white cut-off shorts, and moved from an indoor yoga room to an outdoor pathway. \\

\textbf{Object to Ignore: \texttt{[Object]}}
\end{tcolorbox}

\begin{tcolorbox}[carboxstyle, title=Modification Text Generation for Car subset]
Based on the two provided images, generate a modification text that describes the changes from the first image to the second. \\

\textbf{Important Context:} \\
The car (model and color) is the same in both images. \\

\textbf{Requirements:} \\
1. The modification text must be written in fluent and natural English, NOT exceeding 25 words. \\
2. Focus exclusively on the most significant and definite changes (e.g., Background / Environment, Viewing Angle, Car's State).
   DO NOT describe the car's model or color, as they are unchanged. \\
3. Avoid any explicit references to the images themselves.
   For example, DO NOT use phrases like ``\,in the first image'' or ``\,in the second picture''. \\
4. Employ diverse expressions. Avoid using repetitive sentence structures or fixed grammatical patterns. \\

\textbf{Examples:} \\
1. Now shown from a low-angle perspective. \\
2. The scene changes to a desert at sunset. \\
3. The car is now viewed from a front angle on a snowy mountain road with its headlights turned on. \\
4. Instead of being parked in a garage, the vehicle is now on a bridge with its driver-side door open.
\end{tcolorbox}

\begin{tcolorbox}[productboxstyle, title=Modification Text Generation for Product subset]
Based on the two provided images, generate a modification text that describes the changes from the first image to the second. \\

\textbf{Important Context:} \\
The product object: \texttt{[Object]} is the same in both images. You are strictly forbidden from mentioning
\end{tcolorbox}

\begin{tcolorbox}[productboxstyle]
this product in your response. Your task is to describe how its presentation has changed. \\

\textbf{Requirements:} \\
1. The modification text must be written in fluent and natural English, NOT exceeding 30 words. \\
2. Focus exclusively on the most significant and definite changes (e.g., Background / Environment, Viewing Angle, State, Packaging, Interaction). \\
3. A specific ``\,Object to Ignore'' is provided below. DO NOT mention this product object or any of its attributes (e.g., color, brand, type) in your response. \\
4. Avoid any explicit references to the images themselves.
   For example, DO NOT use phrases like ``\,in the first image'' or ``\,in the second picture''. \\
5. Employ diverse expressions. Avoid using repetitive sentence structures or fixed grammatical patterns. \\

\textbf{Examples:} \\
1. Now shown from a top-down perspective. \\
2. Now shown out of its original packaging. \\
3. The laptop is open and displayed on a wooden desk. \\
4. The sneakers are now being worn by a person on a basketball court. \\

\textbf{Object to Ignore: \texttt{[Object]}}
\end{tcolorbox}

\begin{tcolorbox}[landmarkboxstyle, title=Modification\! Text Generation for Landmark subset]
Based on the two provided images, generate a modification text to transform the first image into the second. \\

\textbf{Important Context:} \\
Both images are about the landmark: \texttt{[Object]}. You are strictly forbidden from mentioning this landmark in your response. 	Your task is to describe how its context, framing, and atmosphere has changed. \\

\textbf{Requirements:} \\
1. The modification text must be written in fluent and natural English, NOT exceeding 30 words. \\
2. Focus exclusively on the most significant and definite changes (e.g., Viewing Angle, Change in Scope or Focus, Atmospheric Conditions, Surrounding Environment). \\
3. A specific ``\,Object to Ignore'' is provided below. DO NOT mention this landmark, its name, its architectural style, or its location in your response. \\
4. Avoid any explicit references to the images themselves.
   For example, DO NOT use phrases like ``\,in the first image'' or ``\,in the second picture''.
\end{tcolorbox}

\begin{tcolorbox}[landmarkboxstyle]
5. Employ diverse expressions. Avoid using repetitive sentence structures or fixed grammatical patterns. \\

\textbf{Examples:} \\
1. Now seen from an aerial perspective on a clear day. \\
2. The scene shifts to a clear night, with the structure illuminated. \\
3. Now viewed from across the river on a foggy morning, with autumn foliage visible. \\

\textbf{Object to Ignore: \texttt{[Object]}}
\end{tcolorbox}

\begin{table}[ht]
    \centering
    \footnotesize
    \setlength{\tabcolsep}{9pt}
    \renewcommand{\arraystretch}{1.05}
    \resizebox{\linewidth}{!}{
        \begin{tabular}{lrr}
        \toprule

        \small{\textbf{Statistic}} & \small{\textbf{Number}} & \small{\textbf{Percentage}} \\

        \midrule

        % \textbf{Total Annotated Quadruples} & \textbf{127,166} & 100.0\% \\
        \textbf{Total Annotated Quadruples} & \textbf{127,166} & \\
        \,\,- \textcolor{deepfashion}{Fashion} & 12,874 & 10.1\% \\
        \,\,- \textcolor{cars}{Car} & 12,728 & 10.0\% \\
        \,\,- \textcolor{products}{Product} & 75,616 & 59.5\% \\
        \,\,- \textcolor{landmarks}{Landmark} & 25,948 & 20.4\% \\

        \midrule

        % \textbf{Total Unique Images} & \textbf{39,495} & 100.0\% \\
        \textbf{Total Unique Images} & \textbf{39,495} & \\
        \,\,- \textcolor{deepfashion}{Fashion} & 1,034 & 2.6\% \\
        \,\,- \textcolor{cars}{Car} & 3,111 & 7.9\% \\
        \,\,- \textcolor{products}{Product} & 27,531 & 69.7\% \\
        \,\,- \textcolor{landmarks}{Landmark} & 7,819 & 19.8\% \\

        \midrule

        % \textbf{Total Unique Instances} & \textbf{2,647} & 100.0\% \\
        \textbf{Total Unique Instances} & \textbf{2,647} & \\
        \,\,- \textcolor{deepfashion}{Fashion} & 80 & 3.0\% \\
        \,\,- \textcolor{cars}{Car} & 199 & 7.5\% \\
        \,\,- \textcolor{products}{Product} & 1,419 & 53.6\% \\
        \,\,- \textcolor{landmarks}{Landmark} & 949 & 35.9\% \\

        \midrule

        Maximum Modification Text Length & 30.0 & - \\
        Average Modification Text Length & 20.2 & - \\

        \bottomrule
        \end{tabular}
    }

    \caption{Statistics of \textbf{OACIRR} Training Dataset.}
    \label{tab:statistics_train}
    \vspace{-2mm}

\end{table}

\begin{table}[ht]
    \centering
    \footnotesize
    \setlength{\tabcolsep}{9pt}
    \renewcommand{\arraystretch}{1.02}
    \resizebox{\linewidth}{!}{
        \begin{tabular}{lrr}
        \toprule

        \small{\textbf{Statistic}} & \small{\textbf{Number}} & \small{\textbf{Percentage}} \\

        \midrule

        % \textbf{Total Annotated Quadruples} & \textbf{33,449} & 100.0\% \\
        \textbf{Total Annotated Quadruples} & \textbf{33,449} & \\
        \,\,- \textcolor{deepfashion}{Fashion} & 3,606 & 10.8\% \\
        \,\,- \textcolor{cars}{Car} & 3,586 & 10.7\% \\
        \,\,- \textcolor{products}{Product} & 21,046 & 62.9\% \\
        \,\,- \textcolor{landmarks}{Landmark} & 5,211 & 15.6\% \\

        \midrule

        % \textbf{Total Unique Images} & \textbf{26,595} & 100.0\% \\
        \textbf{Total Unique Images} & \textbf{26,595} & \\
        Quadruple Images & 15,467 & 58.1\% \\
        Distractor Images & 11,134 & 41.9\% \\
        \,\,- \textcolor{deepfashion}{Fashion} & 5,077 & 19.1\% \\
        \,\,- \textcolor{cars}{Car} & 4,717 & 17.7\% \\
        \,\,- \textcolor{products}{Product} & 11,801 & 44.4\% \\
        \,\,- \textcolor{landmarks}{Landmark} & 5,000 & 18.8\% \\

        \midrule

        % \textbf{Total Unique Instances} & \textbf{4,945} & 100.0\% \\
        \textbf{Total Unique Instances} & \textbf{4,945} & \\
        Quadruple Instances & 1,238 & 25.0\% \\
        Distractor Instances & 3,707 & 75.0\% \\
        \,\,- \textcolor{deepfashion}{Fashion} & 1,683 & 34.0\% \\
        \,\,- \textcolor{cars}{Car} & 1,089 & 22.0\% \\
        \,\,- \textcolor{products}{Product} & 799 & 16.2\% \\
        \,\,- \textcolor{landmarks}{Landmark} & 1,374 & 27.8\% \\

        \midrule

        Maximum Modification Text Length & 30.0 & - \\
        Average Modification Text Length & 19.4 & - \\

        \bottomrule
        \end{tabular}
    }

    \caption{Statistics of \textbf{OACIRR} Evaluation Benchmark.}
    \label{tab:statistics_eval}
    \vspace{-3.1mm}

\end{table}

\subsection{Detailed Dataset Statistics}
\label{subsec:detailed_dataset_statistics}

As shown in Tables~\ref{tab:statistics_train} and~\ref{tab:statistics_eval}, we provide a detailed statistical breakdown of the \textbf{OACIRR} benchmark, highlighting the scale and diversity of both the training data and the evaluation benchmark.
The partitioning and design of \textbf{OACIRR} were guided by two principles to ensure rigor and utility:
\begin{itemize}
\item
    \textbf{Strict data partitioning for fair evaluation.}
    We enforce a strict separation between the training and evaluation splits by ensuring that no images or instances overlap between them. We further reduce fine-grained category overlap to prevent data leakage and ensure that evaluation faithfully reflects generalization to unseen instances.
\item
    \textbf{Asymmetric design for comprehensive evaluation.}
    The asymmetric composition of the four subsets is a deliberate design choice that leverages domain-specific characteristics to assess complementary retrieval capabilities.
    The \textcolor{deepfashion}{Fashion}, \textcolor{cars}{Car}, and \textcolor{landmarks}{Landmark} subsets emphasize \textit{retrieval depth}, requiring discrimination among visually similar instances within a coherent domain.
    In contrast, the \textcolor{products}{Product} subset targets \textit{retrieval breadth}, evaluating robustness under substantially larger and more diverse candidate spaces.
    Collectively, these complementary settings provide a holistic assessment of both fine-grained discrimination and large-scale retrieval performance.
\end{itemize}

\subsection{Instance Diversity Visualization}
\label{subsec:instance_diversity_visualization}

\begin{figure*}[t]
    \centering
        \includegraphics[width=\linewidth]{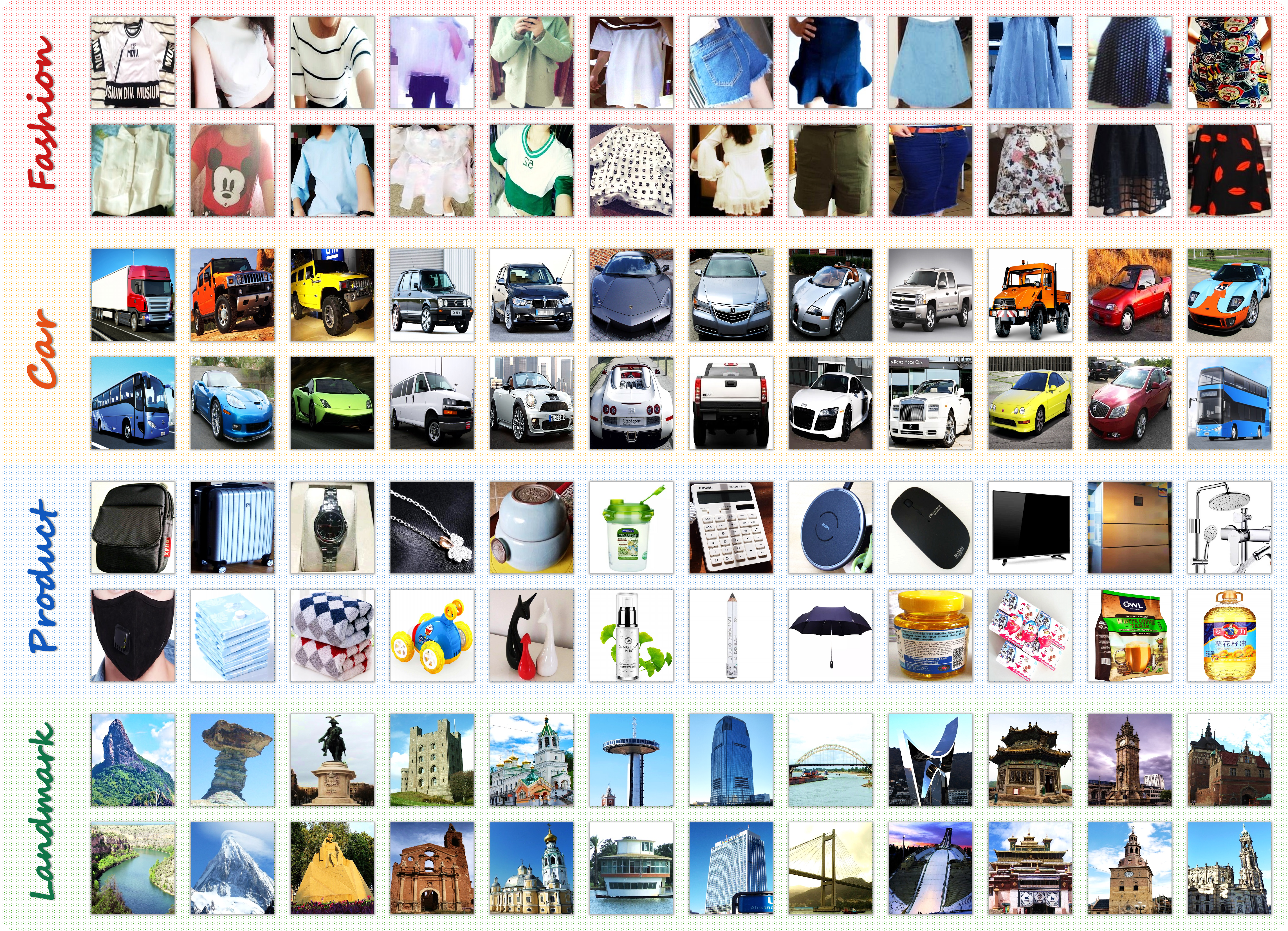}
    \caption{A curated collage of representative instances from the \textbf{OACIRR} benchmark.}
    \label{fig:instance_collage}
\end{figure*}

Figure~\ref{fig:instance_collage} presents a curated collage of representative, cropped instances from the four primary domains, offering a compact visual summary of the benchmark's scope.
\textbf{OACIRR} covers a broad spectrum of categories, ranging from everyday apparel and common vehicles to diverse consumer goods and iconic global sites, exposing models to a wide variety of visual concepts and real-world contexts.

Complementing this breadth, \textbf{OACIRR} also exhibits substantial fine-grained depth.
Individual sub-categories are densely populated with numerous distinct instances, encompassing a wide range of appearance variations.
Such granularity enables evaluation to extend beyond coarse category recognition toward precise, instance-level discrimination.
Collectively, this diversity and depth establish \textbf{OACIRR} as a comprehensive and challenging benchmark for instance-aware compositional retrieval.

\section{Additional Evaluation Protocols and Results}
\label{sec:additional_evaluation_protocols_and_results}

% To supplement the quantitative results in Section~\ref{subsec:quantitative_evaluation}, this section provides the detailed evaluation protocols used to adapt existing retrieval paradigms to the OACIR task and presents additional results under alternative configurations.
To supplement the quantitative results in the main text, this section provides the detailed evaluation protocols used to adapt existing retrieval paradigms to the OACIR task and presents additional results under alternative configurations.
Section~\ref{subsec:details_on_evaluation_protocols} details the two adaptation settings that convert the anchored-instance constraint into formats compatible with different model architectures, and Section~\ref{subsec:ablation_on_evaluation_protocols} reports supplementary quantitative results under these settings.

\subsection{Details on Evaluation Protocols}
\label{subsec:details_on_evaluation_protocols}

\noindent
\textbf{Setting 1: Instance-as-Textual Adaptation.}
The anchored object is specified through a textual cue.
A short template containing the instance's class label is appended to the original modification text, converting the OACIR task into an instance-aware CIR formulation while preserving richer contextual information.
This setting assesses the model's capacity to ground fine-grained textual constraints within a visually complex query.
Prompt templates are given below:

\begin{tcolorbox}[defaultboxstyle, title=Prompt Templates for Setting 1]
\center
% \small
\textbf{1.} \,\,Same \texttt{[Object]} \\
\textbf{2.} \,\,With the same \texttt{[Object]} \\
\textbf{3.} \,\,Fixed \texttt{[Object]} \\
\textbf{4.} \,\,Identical \texttt{[Object]} \\
\textbf{5.} \,\,Invariant \texttt{[Object]} \\
\textbf{6.} \,\,Keep the \texttt{[Object]} \\
\textbf{7.} \,\,Preserving the \texttt{[Object]} \\
\textbf{8.} \,\,\texttt{[Object]} unchanged
\end{tcolorbox}

\noindent
\textbf{Setting 2: Instance-as-Visual Adaptation.}
The anchored object is provided as an explicit visual cue by rendering its bounding box onto the reference image and pairing it with a brief instruction.
This setting assesses the model's capacity to interpret direct visual grounding signals for instance preservation.
The instruction is given below:

\begin{tcolorbox}[defaultboxstyle, title=Instruction for Setting 2]
\center
% \small
% 1. Same \texttt{[Object]} in the \texttt{[Color]} bounding box. \\
% 2. With the same \texttt{[Object]} in the \texttt{[Color]} bounding box. \\
% 3. Fixed \texttt{[Object]} in the \texttt{[Color]} bounding box. \\
% 4. Identical \texttt{[Object]} in the \texttt{[Color]} bounding box. \\
% 5. Invariant \texttt{[Object]} in the \texttt{[Color]} bounding box. \\
% 6. Keep the \texttt{[Object]} in the \texttt{[Color]} bounding box. \\
% 7. Preserving the \texttt{[Object]} in the \texttt{[Color]} bounding box. \\
% 8. \texttt{[Object]} unchanged in the \texttt{[Color]} bounding box.
\texttt{[Prompt Template for Setting 1]} \\
in the \texttt{[Color]} bounding box.
\end{tcolorbox}

\noindent
\textbf{Model-Specific Application.}
Universal Multimodal Retrieval (UMR) models rely heavily on visual grounding and instructional prompts.
Therefore, we adopt \textbf{Setting 2} as the default protocol for these models, using domain–specific instructions tailored to each OACIRR subset.
The complete domain-specific instruction templates are provided below.

\begin{tcolorbox}[fashionboxstyle, title=UMR Instruction Templates for Fashion subset]
\center
\textbf{1.} \,Find a fashion image that aligns with \\
            the reference image and style note. \\
\vspace{2mm}
\textbf{2.} \,Retrieve a fashion scene image that reflects the described transformation from the provided image. \\
\vspace{2mm}
\textbf{3.} \,Can you find an outfit image that meets the adjustments described in the text? \\
\vspace{2mm}
\textbf{4.} \,I'm looking for a similar fashion image with the described changes to the model and scene.
\end{tcolorbox}

\begin{tcolorbox}[carboxstyle, title=UMR Instruction Templates for Car subset]
\center
\textbf{1.} \,Retrieve a car image that aligns with \\
              the reference image and the scene modifications. \\
\vspace{2mm}
\textbf{2.} \,Find a vehicle image like this one, \\
              but with the adjustments from the text. \\
\vspace{2mm}
\textbf{3.} \,Can you pull up a car image that \\
              incorporates the requested changes? \\
\vspace{2mm}
\textbf{4.} \,I'm looking for a similar car image with the described changes to the setting and angle.
\end{tcolorbox}

\begin{tcolorbox}[productboxstyle, title=UMR Instruction Templates for Product subset]
\center
\textbf{1.} \,Find a product image that aligns with \\
              the provided image and the modification instructions. \\
\vspace{2mm}
\textbf{2.} \,Given the reference image and display notes, \\
              find the matching product image. \\
\vspace{2mm}
\textbf{3.} \,Can you find a product image that meets the requested changes to the background and view? \\
\vspace{2mm}
\textbf{4.} \,I'm looking for a similar product image matches \\
              the new display style from the text.
\end{tcolorbox}

\begin{tcolorbox}[landmarkboxstyle, title=UMR Instruction Templates for Landmark subset]
\center
\textbf{1.} \,Retrieve a landmark image that aligns with \\
              the reference image and the described conditions. \\
\vspace{2mm}
\textbf{2.} \,Pull up a photo of a landmark that matches the reference image and the requested transformation. \\
\vspace{2mm}
\textbf{3.} \,Given the reference image and description, \\
              identify the corresponding landmark view. \\
\vspace{2mm}
\textbf{4.} \,I'm looking for a similar landmark image with the specified changes in atmosphere and perspective.
\end{tcolorbox}

In contrast, Zero-shot and Supervised CIR methods do not support bounding-box inputs.
Therefore, we adopt \textbf{Setting 1} as their default protocol, translating the instance constraint into a textual form compatible with their workflow.

\subsection{Ablation on Evaluation Protocols}
\label{subsec:ablation_on_evaluation_protocols}

To validate these choices, we additionally evaluate UMR models under \textit{Setting 1} and CIR models under \textit{Setting 2}.
As shown in Table~\ref{tab:ablation_oacirr_benchmark}, each model class performs best under its default protocol, indicating that UMR models rely on explicit visual grounding while CIR models favor semantically integrated textual cues.
In contrast, our \textbf{\textit{AdaFocal}} provides a robust encoding mechanism that adapts reliably to the OACIR task and its anchored-instance constraint.

\begin{table*}[t]
    \vspace{-2.5mm}
    \centering
    \setlength{\tabcolsep}{5pt}
    \renewcommand{\arraystretch}{1.2}
    \resizebox{\textwidth}{!}{
        \begin{tabular}{c|c|c|ccc|ccc|ccc|ccc|c}
        \toprule

        \multirow{2.4}{*}{\large\textbf{Domain}} & \multirow{2.4}{*}{\large\textbf{Method}} & \multirow{2.4}{*}{\textbf{Pretraining Data}} & \multicolumn{3}{c|}{\large\textcolor{deepfashion}{\textbf{Fashion}}} & \multicolumn{3}{c|}{\large\textcolor{cars}{\textbf{Car}}} & \multicolumn{3}{c|}{\large\textcolor{products}{\textbf{Product}}} & \multicolumn{3}{c|}{\large\textcolor{landmarks}{\textbf{Landmark}}} & \multirow{2.4}{*}{\large\textbf{\textit{Avg.}}} \\

        \cmidrule(lr){4-6} \cmidrule(lr){7-9} \cmidrule(lr){10-12} \cmidrule(lr){13-15}

        &  &  & \textbf{$\text{R}_{\text{ID}}\!\text{@1}$} & \textbf{R@1} & \textbf{R@5} & \textbf{$\text{R}_{\text{ID}}\!\text{@1}$} & \textbf{R@1} & \textbf{R@5} & \textbf{$\text{R}_{\text{ID}}\!\text{@1}$} & \textbf{R@1} & \textbf{R@5} & \textbf{$\text{R}_{\text{ID}}\!\text{@1}$} & \textbf{R@1} & \textbf{R@5} & \\

        \midrule

        \rowcolor{gray!20}
        \multicolumn{16}{c}{\large\textbf{Setting 1: Instance-as-Textual Adaptation}} \\

        \midrule

        \multirow{5}{*}{UMR}
        & LamRA-Ret~\cite{lamra} & M-BEIR + NLI & 25.93 & 20.54 & 36.26 & 58.13 & 33.87 & 72.10 & 67.27 & 36.64 & 67.51 & 57.05 & 32.06 & 67.99 & 47.95 \\
        & MM-Embed~\cite{mm_embed} & M-BEIR + MTEB & 38.05 & 32.70 & 50.69 & 51.37 & 29.62 & 61.74 & 66.68 & 36.73 & 65.49 & 75.95 & 37.75 & 78.53 & 52.11 \\
        & GME (\,2B\,)~\cite{gme} &\multirow{2}{*}{UMRB} & 37.10 & 31.45 & 51.33 & 55.91 & 30.37 & 63.94 & 75.91 & 40.90 & 72.39 & 72.65 & 38.76 & 74.46 & 53.76 \\
        & GME (\,7B\,)~\cite{gme} &  & 44.54 & 38.33 & 59.51 & 58.73 & 35.05 & 70.91 & 81.87 & 53.42 & 82.97 & 76.20 & 46.82 & 82.27 & 60.89 \\
        & U-MARVEL~\cite{u_marvel} & M-BEIR + NLI & 44.32 & 39.14 & 59.64 & 59.63 & 38.17 & 72.16 & 80.78 & 51.40 & 81.01 & 68.00 & 37.08 & 72.23 & 58.63 \\

        \midrule

        \multirow{2}{*}{\textbf{ZS-CIR}}
        & Pic2Word~\cite{pic2word} & \multirow{2}{*}{CC3M} & 14.98 & 11.15 & 21.55 & 12.07 & 4.07 & 11.32 & 45.95 & 13.66 & 34.19 & 55.98 & 20.99 & 52.12 & 24.84 \\
        & LinCIR~\cite{lincir} &  & 15.78 & 12.04 & 21.82 & 5.55 & 2.23 & 7.28 & 47.55 & 14.63 & 34.91 & 42.76 & 19.57 & 47.15 & 22.61 \\

        \midrule

        \multirow{2}{*}{\textbf{CIR}}
        & \multirow{2}{*}{SPRC (ViT-G)~\cite{sprc}} & CIRR & 28.62 & 25.79 & 44.48 & 25.13 & 15.92 & 37.06 & 54.39 & 34.85 & 62.31 & 40.41 & 26.29 & 52.39 & 37.30 \\
        &  & \textbf{OACIRR (Ours)} & 65.25 & 58.51 & 80.89 & 72.87 & 49.82 & 89.57 & 86.05 & 70.61 & 93.68 & 76.32 & \underline{56.04} & 89.00 & 74.05 \\

        \midrule

        \rowcolor{gray!20}
        \multicolumn{16}{c}{\large\textbf{Setting 2: Instance-as-Visual Adaptation}} \\

        \midrule

        \multirow{5}{*}{\textbf{UMR}}
        & LamRA-Ret~\cite{lamra} & M-BEIR + NLI & 27.45 & 21.63 & 37.10 & 61.03 & 35.44 & 74.51 & 69.45 & 39.53 & 70.25 & 58.64 & 32.58 & 68.74 & 49.70 \\
        & MM-Embed~\cite{mm_embed} & M-BEIR + MTEB & 41.38 & 34.55 & 52.50 & 53.21 & 30.06 & 62.80 & 71.03 & 41.47 & 71.15 & 78.85 & 38.88 & 79.32 & 54.60 \\
        & GME (\,2B\,)~\cite{gme} &\multirow{2}{*}{UMRB} & 38.13 & 32.14 & 51.50 & 58.84 & 31.60 & 66.03 & 76.89 & 44.11 & 74.20 & 73.86 & 38.99 & 75.61 & 55.16 \\
        & GME (\,7B\,)~\cite{gme} &  & 44.98 & 39.24 & 60.18 & 63.11 & 38.34 & 75.38 & 83.44 & 54.60 & 84.15 & 77.11 & 47.09 & 82.69 & 62.53 \\
        & U-MARVEL~\cite{u_marvel} & M-BEIR + NLI & 46.05 & 40.38 & 60.59 & 62.92 & 39.96 & 74.90 & 83.26 & 54.69 & 84.13 & 69.81 & 37.67 & 73.08 & 60.62 \\

        \midrule

        \multirow{2}{*}{ZS-CIR}
        & Pic2Word~\cite{pic2word} & \multirow{2}{*}{CC3M} & 14.96 & 11.00 & 21.16 & 12.04 & 3.95 & 11.07 & 39.39 & 11.50 & 27.13 & 46.60 & 18.39 & 46.49 & 21.97 \\
        & LinCIR~\cite{lincir} &  & 15.76 & 11.99 & 21.48 & 5.54 & 2.17 & 7.25 & 46.57 & 13.85 & 33.96 & 42.16 & 19.09 & 47.11 & 22.24 \\

        \midrule

        \multirow{2}{*}{CIR}
        & \multirow{2}{*}{SPRC (ViT-G)~\cite{sprc}} & CIRR & 28.59 & 25.68 & 43.68 & 24.23 & 15.48 & 36.25 & 46.62 & 29.33 & 49.44 & 33.64 & 23.03 & 46.73 & 33.56 \\
        &  & \textbf{OACIRR (Ours)} & 64.14 & 57.71 & 79.65 & 72.70 & 48.29 & 89.18 & 84.27 & 66.86 & 91.13 & 75.24 & 54.65 & 88.93 & 72.73 \\

        \midrule

        \rowcolor{gray!20}
        \multicolumn{16}{c}{\large\textbf{OACIR Task-Specific Architecture}} \\

        \midrule

        \rowcolor{gray!5}
        & \textbf{Baseline (ViT-G)} &  & 69.07 & 58.76 & 81.44 & 74.59 & 49.78 & 89.46 & 87.48 & 69.53 & 93.66 & 79.80 & 55.49 & 89.87 & 74.91 \\
        \rowcolor{gray!5}
        & \textbf{$\text{Baseline}^{\dagger}$ (ViT-G)} &  & \underline{72.66} & \underline{63.31} & \underline{83.97} & \underline{76.85} & 50.24 & \underline{89.87} & \underline{88.68} & \underline{72.13} & \underline{94.09} & \underline{80.05} & 55.69 & \underline{90.14} & \underline{76.47} \\
        \rowcolor{gray!5}
        & \textbf{$\text{SPRC}^{\dagger}$ (ViT-G)} &  & 69.94 & 60.98 & 82.72 & 74.08 & \underline{51.62} & 89.79 & 86.42 & 70.90 & 93.74 & 77.41 & 55.90 & 89.02 & 75.21 \\

        \rowcolor{orange!10}
        \multirow{-4}{*}{\textbf{OACIR}} & \textbf{\textit{AdaFocal} (ViT-G)} & \multirow{-4}{*}{\textbf{OACIRR (Ours)}} & \textbf{77.15} & \textbf{65.31} & \textbf{86.88} & \textbf{78.42} & \textbf{53.63} & \textbf{92.22} & \textbf{91.86} & \textbf{74.11} & \textbf{95.39} & \textbf{82.92} & \textbf{58.47} & \textbf{91.63} & \textbf{79.00} \\

        \bottomrule
        \end{tabular}
    }

    \caption{\textbf{Quantitative comparison on the OACIRR benchmark under different evaluation settings and OACIR-specific baselines.} ``Avg.'' represents the average results across all evaluation metrics. The best result is highlighted in \textbf{bold}, and the second best is \underline{underlined}. \textbf{Baseline} denotes the standard CIR baseline, \textbf{$\text{Baseline}^{\dagger}$} denotes the ROI-cropped baseline, and \textbf{$\text{SPRC}^{\dagger}$} denotes plug-and-play CIR baseline.}
    \label{tab:ablation_oacirr_benchmark}
    \vspace{-1.5mm}
\end{table*}

\section{Additional Ablation Studies}
\label{sec:additional_ablation_studies}

This section provides additional ablation studies to further validate the effectiveness of our method and the value of the \textbf{OACIRR} benchmark.
Section~\ref{subsec:region_aware_baseline} compares \textbf{\textit{AdaFocal}} against stronger region-aware baselines, including explicit ROI cropping and a plug-and-play integration into SPRC~\cite{sprc}, further verifying the effectiveness of our adaptive attention design.
Section~\ref{subsec:cross_task_generalization_of_oacirr} and Section~\ref{subsec:cross_domain_generalization_on_oacirr} evaluate the generalization ability of models trained on \textbf{OACIRR} across tasks and domains, respectively.
Section~\ref{subsec:robustness_to_bounding_box_quality} analyzes the robustness of \textbf{\textit{AdaFocal}} to imperfect bounding box inputs.
Finally, Section~\ref{subsec:caam_design_analysis} examines key design choices within the \textit{Context-Aware Attention Modulator (CAAM)}, including the modulation output form, the number of self-attention layers, and the configuration of contextual probe tokens.

\begin{table*}[t]
    \centering
    \setlength{\tabcolsep}{6pt}
    \renewcommand{\arraystretch}{1.25}
    \resizebox{\textwidth}{!}{
        \begin{tabular}{c|c|r|cc|cccc|ccc}
        \toprule

        \multirow{2.4}{*}{\large\textbf{Method}} & \multirow{2.4}{*}{\textbf{Pretraining Data}} & \multirow{2.4}{*}{\textbf{Pretraining Scale}} & \multicolumn{2}{c|}{\textbf{FashionIQ}} & \multicolumn{4}{c|}{\textbf{CIRR}} & \multicolumn{3}{c}{\textbf{CIRCO}} \\

        \cmidrule(lr){4-5} \cmidrule(lr){6-9} \cmidrule(lr){10-12}

        &  &  & \textbf{Avg@10} & \textbf{Avg@50} & \textbf{R@1} & \textbf{R@5} & \textbf{$\text{R}_{\text{s}}\!\text{@1}$} & \textbf{Avg.} & \textbf{mAP@5} & \textbf{mAP@10} & \textbf{mAP@25} \\

        \midrule

        CASE~\cite{data_roaming} & LaSCo + CoCo & 389\,K\,\,\,\,\,\,\,\,\,\,\,\,\, & -- & -- & 35.40 & 65.78 & 64.29 & 65.04 & -- & -- & -- \\
        CoVR-BLIP~\cite{covr} & WebVid-CoVR & 1,644\,K\,\,\,\,\,\,\,\,\,\,\,\,\, & 27.70 & 44.63 & 38.48 & 66.70 & 69.28 & 67.99 & 21.43 & 22.33 & 24.47 \\
        CompoDiff~\cite{compodiff} & ST18M + LAION-2B & 18,000\,K\,\,\,\,\,\,\,\,\,\,\,\,\, & 39.02 & 51.71 & 26.71 & 55.14 & 64.54 & 59.84 & 15.33 & 17.71 & 19.45 \\
        CoAlign (ViT-G)~\cite{cirhs} & CIRHS & 535\,K\,\,\,\,\,\,\,\,\,\,\,\,\, & \underline{39.22} & \underline{60.08} & 41.08 & 71.11 & 70.80 & 70.96 & \underline{21.60} & \underline{23.38} & \underline{25.98} \\

        \midrule

        \rowcolor{gray!10}
        \textbf{Baseline (ViT-L)} &  &  & 37.82 & 59.57 & \underline{41.59} & \underline{72.36} & \underline{72.06} & \underline{72.21} & 21.51 & 23.14 & 25.47 \\
        \rowcolor{orange!10}
        \textbf{Baseline (ViT-G)} & \multirow{-2}{*}{\textbf{OACIRR (Ours)}} & \multirow{-2}{*}{\textbf{127\,K\,\,\,\,\,\,\,\,\,\,\,\,\,}} & \textbf{39.80} & \textbf{61.81} & \textbf{42.96} & \textbf{73.62} & \textbf{72.95} & \textbf{73.28} & \textbf{23.96} & \textbf{24.69} & \textbf{26.58} \\

        \bottomrule
        \end{tabular}
    }

    \caption{Zero-shot cross-task generalization on standard CIR benchmarks.}
    \label{tab:cross_task_generalization}
\end{table*}

\begin{table*}[t]
    \centering
    \setlength{\tabcolsep}{8.8pt}
    \renewcommand{\arraystretch}{1.25}
    \resizebox{\linewidth}{!}{
        \begin{tabular}{c|c|ccc|ccc|ccc|ccc|c}
        \toprule

        \multirow{2.4}{*}{\large\textbf{Setting}} & \multirow{2.4}{*}{\textbf{Method}} & \multicolumn{3}{c|}{\large\textcolor{deepfashion}{\textbf{Fashion}}} & \multicolumn{3}{c|}{\large\textcolor{cars}{\textbf{Car}}} & \multicolumn{3}{c|}{\large\textcolor{products}{\textbf{Product}}} & \multicolumn{3}{c|}{\large\textcolor{landmarks}{\textbf{Landmark}}} & \multirow{2.4}{*}{\large\textbf{\textit{Avg.}}} \\

        \cmidrule(lr){3-5} \cmidrule(lr){6-8} \cmidrule(lr){9-11} \cmidrule(lr){12-14}

        &  & \textbf{$\text{R}_{\text{ID}}\!\text{@1}$} & \textbf{R@1} & \textbf{R@5} & \textbf{$\text{R}_{\text{ID}}\!\text{@1}$} & \textbf{R@1} & \textbf{R@5} & \textbf{$\text{R}_{\text{ID}}\!\text{@1}$} & \textbf{R@1} & \textbf{R@5} & \textbf{$\text{R}_{\text{ID}}\!\text{@1}$} & \textbf{R@1} & \textbf{R@5} & \\

        \midrule

        & SPRC~\cite{sprc} & 48.73 & 40.51 & 62.84 & 66.86 & 41.49 & 78.98 & 62.79 & 42.68 & 71.72 & 59.49 & 38.55 & 71.23 & 57.16 \\
        \rowcolor{gray!10}
        \multirow{-2}{*}{\textbf{Cross-Domain}} & \textbf{\textit{AdaFocal}} & \textbf{61.15} & \textbf{50.25} & \textbf{71.54} & \textbf{74.26} & \textbf{45.65} & \textbf{85.81} & \textbf{67.84} & \textbf{45.53} & \textbf{74.68} & \textbf{61.58} & \textbf{40.66} & \textbf{71.94} & \textbf{62.57} \\

        \midrule

        & SPRC~\cite{sprc} & 65.25 & 58.51 & 80.89 & 72.87 & 49.82 & 89.57 & 86.05 & 70.61 & 93.68 & 76.32 & 56.04 & 89.00 & 74.05 \\
        \rowcolor{orange!10}
        \multirow{-2}{*}{\textbf{Full Finetuning}} & \textbf{\textit{AdaFocal}} & \textbf{77.15} & \textbf{65.31} & \textbf{86.88} & \textbf{78.42} & \textbf{53.63} & \textbf{92.22} & \textbf{91.86} & \textbf{74.11} & \textbf{95.39} & \textbf{82.92} & \textbf{58.47} & \textbf{91.63} & \textbf{79.00} \\

        \bottomrule
        \end{tabular}
    }

    \caption{Cross-domain generalization on the \textbf{OACIRR} benchmark.}
    % ``Cross-Domain'' denotes leave-one-domain-out evaluation, where models are trained on three subsets and tested on the held-out one. ``Full Finetuning'' denotes training on all four subsets.
    \label{tab:cross_domain_generalization}
\end{table*}

\subsection{Region-Aware Baselines}
\label{subsec:region_aware_baseline}

To rigorously evaluate the necessity and effectiveness of our adaptive attention mechanism, we compare \textbf{\textit{AdaFocal}} against three region-aware baseline models, each reflecting a distinct way of handling the anchored instance constraint:
\begin{itemize}
\item
    \textbf{Standard CIR Baseline.}
    This model removes the \textit{CAAM} module ($\beta=0$) and encodes the full reference image and modification text using the Multimodal Encoder. While it preserves global visual context, it lacks any mechanism to preferentially attend to the anchored instance region.
\item
    \textbf{ROI-Cropped Baseline ($\text{Baseline}^{\dagger}$).}
    To introduce explicit region awareness without additional learning, we crop the reference image using the bounding box $B_r$ and feed only the cropped region into the encoder. This forces attention onto the instance but eliminates surrounding context essential for interpreting the modification text.
\item
    \textbf{Plug-and-Play CIR Baseline ($\text{SPRC}^{\dagger}$).}
    To enable a fairer independent evaluation, we integrate the \textit{CAAM} into the strong CIR model SPRC~\cite{sprc} by applying its dynamic attention activation during the first image-text fusion stage of the query encoder. This isolates \textit{CAAM} as a plug-and-play module for instance-focused attention modulation within an existing CIR architecture.
\end{itemize}

As shown in Table~\ref{tab:ablation_oacirr_benchmark}, using the cropped instance ($\text{Baseline}^{\dagger}$) improves Instance Recall over the Standard Baseline, indicating that explicit isolation strengthens identity preservation.
However, its gains in standard Recall remain limited, suggesting that removing background context hinders the interpretation of contextual modifications.
Integrating \textit{CAAM} into SPRC ($\text{SPRC}^{\dagger}$) still improves over vanilla SPRC, verifying that our module is effective as a plug-and-play instance-aware attention mechanism.
However, its gains remain limited, indicating that complex post-interaction layers prior to query encoding in existing CIR methods can dilute this direct instance-focused attention.
In contrast, \textbf{\textit{AdaFocal}} achieves the strongest overall balance between instance fidelity and compositional reasoning.

\subsection{Cross-Task Generalization of OACIRR}
\label{subsec:cross_task_generalization_of_oacirr}

We evaluate whether the instance-consistent supervision provided by \textbf{OACIRR} transfers effectively to standard CIR settings.
To this end, we train a Standard CIR Baseline exclusively on the \textbf{OACIRR} training set and directly evaluate the resulting model in a zero-shot manner on three established CIR benchmarks: FashionIQ~\cite{fashioniq}, CIRR~\cite{cirr}, and CIRCO~\cite{searle}.
We compare its performance with representative CIR models trained on large-scale or synthetic triplet datasets, including CASE~\cite{data_roaming}, CoVR-BLIP~\cite{covr}, CompoDiff~\cite{compodiff}, and CoAlign~\cite{cirhs}.

As shown in Table~\ref{tab:cross_task_generalization}, the model pretrained on \textbf{OACIRR} achieves strong zero-shot transfer performance across all three benchmarks, consistently outperforming methods trained on substantially larger datasets.
These findings support two key conclusions:
\textbf{(1) \textit{Importance of Instance-Consistent Supervision}}: Enforcing precise instance-level alignment provides a more reliable training signal than synthetic or loosely paired semantic triplets, fostering robust compositional reasoning.
\textbf{(2) \textit{Data Efficiency through High Quality}}: The real-world fidelity and careful curation of \textbf{OACIRR} lead to highly competitive transfer performance while requiring substantially fewer training samples than existing large-scale datasets.
Overall, these cross-task results demonstrate that \textbf{OACIRR} serves not only as a rigorous benchmark for instance-aware retrieval, but also as an effective pretraining resource for the standard CIR task.

\subsection{Cross-Domain Generalization on OACIRR}
\label{subsec:cross_domain_generalization_on_oacirr}

To evaluate whether models trained on \textbf{OACIRR} can generalize beyond domain-specific semantics, we conduct a leave-one-domain-out evaluation across the four subsets.
For each target subset, the model is trained on the remaining three subsets and tested on the held-out one.
We compare this \textbf{Cross-Domain} setting with the standard \textbf{Full Finetuning} setting, where all four subsets are used for training.

As shown in Table~\ref{tab:cross_domain_generalization}, \textbf{\textit{AdaFocal}} consistently outperforms SPRC on all unseen domains under the Cross-Domain setting, demonstrating stronger instance-centric reasoning beyond domain-specific semantics.
At the same time, the clear performance gap between Cross-Domain and Full Finetuning confirms that the four subsets are strongly complementary rather than redundant, highlighting both the diversity and the intrinsic challenge of the \textbf{OACIRR} benchmark.

\begin{table*}[t]
    \centering
    \setlength{\tabcolsep}{8pt}
    \renewcommand{\arraystretch}{1.25}
    \resizebox{\linewidth}{!}{
        \begin{tabular}{cc|ccc|ccc|ccc|ccc|c}
        \toprule

        \multicolumn{2}{c|}{\large\textbf{\,\,\,\,Bounding Box\,\,\,}}& \multicolumn{3}{c|}{\large\textcolor{deepfashion}{\textbf{Fashion}}} & \multicolumn{3}{c|}{\large\textcolor{cars}{\textbf{Car}}} & \multicolumn{3}{c|}{\large\textcolor{products}{\textbf{Product}}} & \multicolumn{3}{c|}{\large\textcolor{landmarks}{\textbf{Landmark}}} & \multirow{2.4}{*}{\large\textbf{\textit{Avg.}}} \\

        \cmidrule(lr){1-2} \cmidrule(lr){3-5} \cmidrule(lr){6-8} \cmidrule(lr){9-11} \cmidrule(lr){12-14}

        \textbf{\,\,\,\,\,\,\,\,\,\,IoU} & \textbf{Perturbation} & \textbf{$\text{R}_{\text{ID}}\!\text{@1}$} & \textbf{R@1} & \textbf{R@5} & \textbf{$\text{R}_{\text{ID}}\!\text{@1}$} & \textbf{R@1} & \textbf{R@5} & \textbf{$\text{R}_{\text{ID}}\!\text{@1}$} & \textbf{R@1} & \textbf{R@5} & \textbf{$\text{R}_{\text{ID}}\!\text{@1}$} & \textbf{R@1} & \textbf{R@5} & \\

        \midrule

        \rowcolor{orange!10}
        \textbf{\,\,\,\,\,\,\,\,\,\,1.00} & \textbf{\textit{Original}} & \textbf{77.15} & \textbf{65.31} & \textbf{86.88} & \textbf{78.42} & \textbf{53.63} & \textbf{92.22} & \textbf{91.86} & \textbf{74.11} & \textbf{95.39} & \textbf{82.92} & \textbf{58.47} & \textbf{91.63} & \textbf{79.00} \\

        \midrule

        \textbf{\,\,\,\,\,\,\,\,\,\,0.80} & \textcolor{products}{\textbf{\textit{Scale}}} & 77.05 & 65.24 & 86.82 & 78.26 & 53.54 & 92.16 & 91.86 & 74.11 & 95.35 & 82.83 & 58.41 & 91.60 & 78.93 \\
        \textbf{\,\,\,\,\,\,\,\,\,\,0.50} & \textbf{\textcolor{products}{\textit{Scale}} + \textcolor{cars}{\textit{Shift}}} & 75.16 & 63.24 & 85.55 & 77.07 & 52.61 & 91.54 & 91.20 & 73.44 & 94.83 & 81.66 & 57.66 & 90.96 & 77.91 \\

        \midrule

        \rowcolor{gray!10}
        \textbf{\,\,\,\,\,\,\,\,\,\,NaN} & \textbf{\textit{w/o Bounding Box}} & 69.07 & 58.76 & 81.44 & 74.59 & 49.78 & 89.46 & 87.48 & 69.53 & 93.66 & 79.80 & 55.49 & 89.87 & 74.91 \\

        \bottomrule
        \end{tabular}
    }

    \vspace{-0.5mm}
    \caption{Robustness of \textbf{\textit{AdaFocal}} to \textcolor{products}{Scale} and \textcolor{cars}{Shift} perturbations of bounding boxes on the \textbf{OACIRR} benchmark.}
    \label{tab:bbox_robustness}
\end{table*}

\begin{table*}[t]
    \centering
    \setlength{\tabcolsep}{10pt}
    \renewcommand{\arraystretch}{1.25}
    \resizebox{\linewidth}{!}{
        \begin{tabular}{c|ccc|ccc|ccc|ccc|c}
        \toprule

        \multirow{2.4}{*}{\textbf{Modulation Output}} & \multicolumn{3}{c|}{\large\textcolor{deepfashion}{\textbf{Fashion}}} & \multicolumn{3}{c|}{\large\textcolor{cars}{\textbf{Car}}} & \multicolumn{3}{c|}{\large\textcolor{products}{\textbf{Product}}} & \multicolumn{3}{c|}{\large\textcolor{landmarks}{\textbf{Landmark}}} & \multirow{2.4}{*}{\large\textbf{\textit{Avg.}}} \\

        \cmidrule(lr){2-4} \cmidrule(lr){5-7} \cmidrule(lr){8-10} \cmidrule(lr){11-13}

        & \textbf{$\text{R}_{\text{ID}}\!\text{@1}$} & \textbf{R@1} & \textbf{R@5} & \textbf{$\text{R}_{\text{ID}}\!\text{@1}$} & \textbf{R@1} & \textbf{R@5} & \textbf{$\text{R}_{\text{ID}}\!\text{@1}$} & \textbf{R@1} & \textbf{R@5} & \textbf{$\text{R}_{\text{ID}}\!\text{@1}$} & \textbf{R@1} & \textbf{R@5} & \\

        \midrule

        \rowcolor{orange!10}
        \textbf{Scalar ( $\beta$ )} & \textbf{77.15} & \textbf{65.31} & \textbf{86.88} & \textbf{78.42} & \textbf{53.63} & \textbf{92.22} & \textbf{91.86} & \textbf{74.11} & \textbf{95.39} & \textbf{82.92} & 58.47 & 91.63 & \textbf{79.00} \\
        \textbf{Vector ( $\vec\beta$ )} & 74.60 & 65.25 & 85.94 & 77.32 & 53.33 & 92.19 & 91.56 & 73.13 & 94.92 & 82.80 & \textbf{58.96} & \textbf{91.77} & 78.48 \\

        \bottomrule
        \end{tabular}
    }

    \vspace{-0.5mm}
    \caption{Ablation study on the modulation output design of the \textit{CAAM}.}
    \label{tab:ablation_caam_output}
\end{table*}

\subsection{Robustness to Bounding Box Quality}
\label{subsec:robustness_to_bounding_box_quality}

To evaluate the robustness of \textbf{\textit{AdaFocal}} to imperfect user inputs, we simulate noisy bounding boxes through \textcolor{products}{Scale} and \textcolor{cars}{Shift} perturbations.
Specifically, \textit{\textcolor{products}{Scale}} enlarges or shrinks the bounding box while preserving its center, and \textit{\textcolor{cars}{Shift}} additionally offsets the center to mimic localization errors.

As shown in Table~\ref{tab:bbox_robustness}, \textbf{\textit{AdaFocal}} is robust to \textcolor{products}{Scale} perturbation, with only negligible performance drops across all subsets.
In contrast, the combined perturbation of \textit{\textcolor{products}{Scale} + \textcolor{cars}{Shift}} causes a clearer degradation, and removing the bounding box leads to the largest drop.
These results indicate that \textbf{\textit{AdaFocal}} tolerates moderate input noise while still relying on visual anchors for reliable instance-aware retrieval.

\begin{table}[t]
    \vspace{-2mm}
    \centering
    \small
    \setlength{\tabcolsep}{9pt}
    \renewcommand{\arraystretch}{1.23}
    \resizebox{\linewidth}{!}{
        \begin{tabular}{c|cccc}
        \toprule

        \textbf{CAAM} & \multicolumn{4}{c}{\textbf{OACIRR Benchmark}} \\
        \cmidrule(lr){1-1} \cmidrule(lr){2-5}
        \textbf{\# Self-Attention Layers} & $\text{R}_{\text{ID}}\!\text{@1}$ & R@1 & R@5 & \textbf{Avg.} \\

        \midrule

        \textbf{1} & 81.38 & 62.39 & 90.54 & 78.10 \\
        \rowcolor{orange!10}
        \textbf{2} & \textbf{82.59} & \textbf{62.88} & \textbf{91.53} & \textbf{79.00} \\
        \textbf{3} & \underline{82.31} & \underline{62.75} & \underline{91.42} & \underline{78.83} \\
        \textbf{4} & 82.02 & 62.51 & 91.24 & 78.59 \\

        \bottomrule
        \end{tabular}
    }

    \vspace{-0.5mm}
    \caption{Ablation study on the number of \textit{self-attention layers}.}
    \label{tab:ablation_caam_self_attention_layer}
    \vspace{-1mm}
\end{table}

\subsection{CAAM Design Analysis}
\label{subsec:caam_design_analysis}

We further analyze three key design choices of the \textit{Context-Aware Attention Modulator (CAAM)}, including the modulation output form, the depth of the \textit{Contextual Reasoning Module (CRM)}, and the number of learnable \textit{Contextual Probe Tokens}.
We prioritize configurations that achieve strong performance with minimal complexity.

\begin{itemize}
\item
    \textbf{Scalar vs. Vector Modulation.}
    As shown in Table~\ref{tab:ablation_caam_output}, replacing the default scalar modulation with a query-wise vector output ($\vec{\beta}\!\in\!\mathbb{R}^{M}$) offers no additional gain. This suggests that a single scalar is sufficient to control attention intensity while better preserving the relative semantic coherence among pre-trained fusion queries. Therefore, we adopt the \textbf{scalar} design as the default output form.
\item
    \textbf{Depth of the Contextual Reasoning Module.}
    As shown in Table~\ref{tab:ablation_caam_self_attention_layer}, increasing the \textit{CRM} depth from 1 to 2 layers leads to clear improvements in both instance-level fidelity and overall recall, indicating that a single layer lacks sufficient cross-modal reasoning capacity. Scaling beyond 2 layers offers no significant gains and may add unnecessary complexity to the compact design of the module. Based on these observations, we employ a \textbf{2-layer CRM}.
\item
    \textbf{Number of the Contextual Probe Tokens.}
    As shown in Table~\ref{tab:ablation_caam_probe_token}, using too few probe tokens limits the module’s capacity to capture diverse contextual cues, while increasing the token count further yields only marginal benefit. Since the performance saturates at \textbf{8 probe tokens}, we adopt this configuration as the default setting.
\end{itemize}

\begin{table}[t]
    \vspace{-2mm}
    \centering
    \small
    \setlength{\tabcolsep}{12pt}
    \renewcommand{\arraystretch}{1.05}
    \resizebox{\linewidth}{!}{
        \begin{tabular}{c|cccc}
        \toprule

        \textbf{CAAM} & \multicolumn{4}{c}{\textbf{OACIRR Benchmark}} \\
        \cmidrule(lr){1-1} \cmidrule(lr){2-5}
        \textbf{\# Probe Tokens} & $\text{R}_{\text{ID}}\!\text{@1}$ & R@1 & R@5 & \textbf{Avg.} \\

        \midrule

        \textbf{2} & 81.92 & \textbf{63.15} & \underline{91.49} & 78.85 \\
        \textbf{4} & 82.38 & 62.41 & 91.15 & 78.65 \\
        \rowcolor{orange!10}
        \textbf{8} & \textbf{82.59} & 62.88 & \textbf{91.53} & \textbf{79.00} \\
        \textbf{16} & \underline{82.46} & \underline{62.94} & 91.45 & \underline{78.95} \\
        \textbf{32} & 82.21 & 62.90 & 91.37 & 78.83 \\

        \bottomrule
        \end{tabular}
    }

    \vspace{-0.5mm}
    \caption{Ablation study on the number of \textit{probe tokens}.}
    \label{tab:ablation_caam_probe_token}
    \vspace{-1mm}
\end{table}

\begin{figure*}[t]
    \centering
    \vspace{2mm}
        \includegraphics[width=\linewidth, trim=0.6cm 0.25cm 0cm 0.5cm, clip]{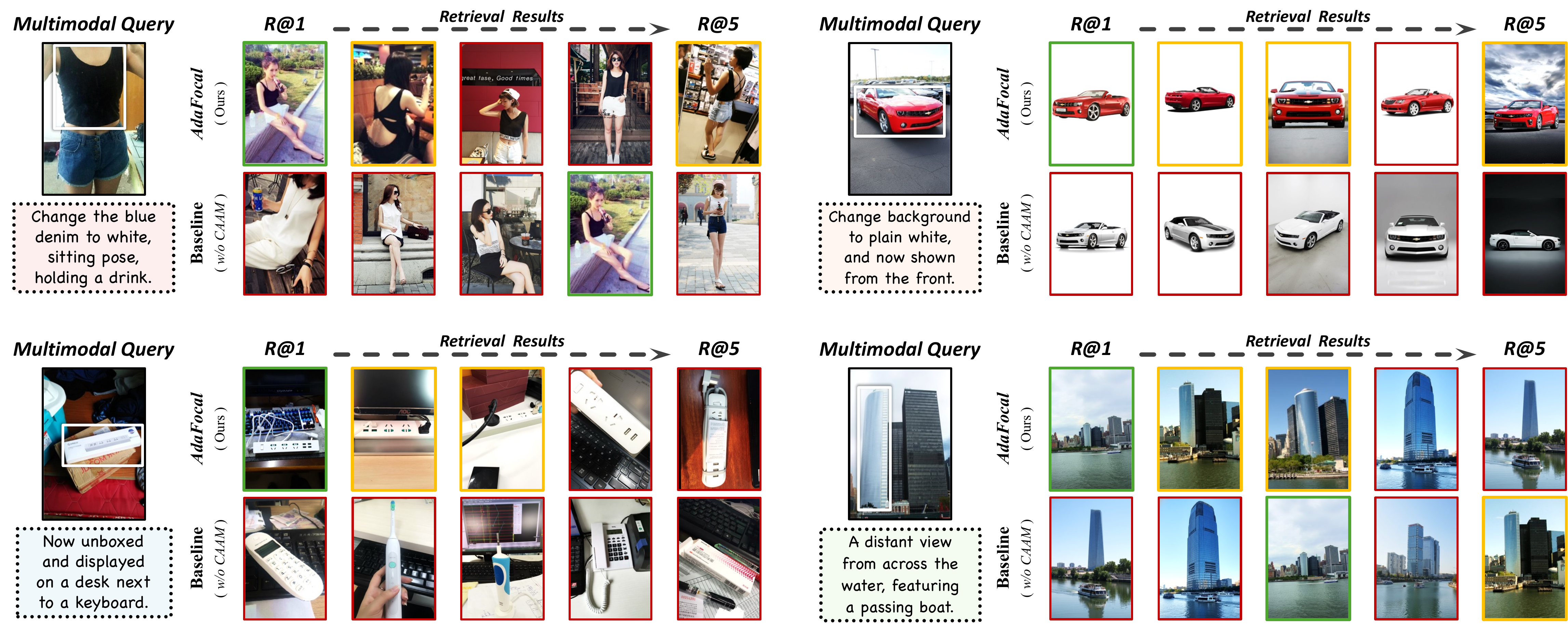}
    \caption{Qualitative comparison of our \textbf{\textit{AdaFocal}} and the Baseline on the \textbf{OACIRR} benchmark. \textbf{\textcolor[HTML]{4EA72E}{Green boxes}} indicate the ground-truth target, \textbf{\textcolor[HTML]{FFC000}{yellow boxes}} indicate instance-correct but semantically incorrect results, and all other retrieved images are marked with \textbf{\textcolor[HTML]{C00000}{red boxes}}.}
    \label{fig:qualitative_comparison_SM}
\end{figure*}

\section{Additional Qualitative Analysis}
\label{sec:additional_qualitative_analysis}

Figure~\ref{fig:qualitative_comparison_SM} presents qualitative comparisons across diverse retrieval scenarios, revealing two failure modes of baseline CIR models and showing how \textbf{\textit{AdaFocal}} addresses them.

\noindent
\textbf{Semantic Drift.}
Baseline models tend to conflate strong textual modifications with intrinsic object attributes, yielding retrievals that follow text-implied properties rather than preserving the visual anchor.
\textbf{\textit{AdaFocal}} maintains instance identity while faithfully reflecting contextual changes.

\noindent
\textbf{Fine-grained Confusion.}
Baseline models often return semantically similar yet instance-incorrect distractors, reflecting a reliance on global semantics over instance-specific cues.
\textbf{\textit{AdaFocal}} retrieves the correct instance more reliably under high visual similarity, offering clear gains in challenging cases by emphasizing distinctive local cues.

\afterpage{\clearpage}

% {
%     \small
%     \bibliographystyle{ieeenat_fullname}
%     \bibliography{OACIR}
% }

\end{document}